\documentclass[10pt,conference]{IEEEtran}

\usepackage{cite}
\usepackage{amsmath,amssymb,amsfonts}
\usepackage{algorithmic}
\usepackage{graphicx}
\usepackage{textcomp}
\usepackage{xcolor}
\usepackage[hyphens]{url}
\usepackage{fancyhdr}
\usepackage{hyperref}
\usepackage[ruled,vlined,linesnumbered]{algorithm2e}
\usepackage{booktabs}
\usepackage{multirow}
\usepackage{tabularx}
\usepackage{subfig}
\newcommand{\hpcayear}{2026}

\newcommand{\hpcasubmissionnumber}{2104}
\title{HA-RAG: Hotness-Aware RAG Acceleration via Mixed Precision and Data Placement}

\def\hpcacameraready{} 
\def\hpcacameraready{} 

\providecommand\hpcaauthors{}   
\providecommand\hpcaaffiliation{}
\providecommand\hpcaemail{}

\renewcommand\hpcaauthors{Danying Ge, Jianhua Gao, Yixue Yang, and Weixing Ji}
\renewcommand\hpcaaffiliation{Beijing Normal University, China}
\renewcommand\hpcaemail{gdanying@163.com, gaojh@bnu.edu.cn, yyxmaeve@gmail.com, jwx@bnu.edu.cn}

\author{
  \ifdefined\hpcacameraready
    \IEEEauthorblockN{\hpcaauthors{}}
      \IEEEauthorblockA{
        \hpcaaffiliation{} \\
        \hpcaemail{}
      }
  \else
    \IEEEauthorblockN{\normalsize{HPCA \hpcayear{} Submission
      \textbf{\#\hpcasubmissionnumber{}}} \\
      \IEEEauthorblockA{
        Confidential Draft \\
        Do NOT Distribute!!
      }
    }
  \fi 
}

\fancypagestyle{camerareadyfirstpage}{%
  \fancyhead{}
  
  \fancyhead[C]{
    \ifdefined\aeopen
    \parbox[][12mm][t]{13.5cm}{\hpcayear{} IEEE International Symposium on High-Performance Computer Architecture (HPCA)}    
    \else
      \ifdefined\aereviewed
      \parbox[][12mm][t]{13.5cm}{\hpcayear{} IEEE International Symposium on High-Performance Computer Architecture (HPCA)}
      \else
      \ifdefined\aereproduced
      \parbox[][12mm][t]{13.5cm}{\hpcayear{} IEEE International Symposium on High-Performance Computer Architecture (HPCA)}
      \else
      \parbox[][0mm][t]{13.5cm}{\hpcayear{} IEEE International Symposium on High-Performance Computer Architecture (HPCA)}
    \fi 
    \fi 
    \fi 
    \ifdefined\aeopen 
      \includegraphics[width=12mm,height=12mm]{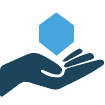}
    \fi 
    \ifdefined\aereviewed
      \includegraphics[width=12mm,height=12mm]{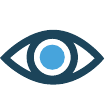}
    \fi 
    \ifdefined\aereproduced
      \includegraphics[width=12mm,height=12mm]{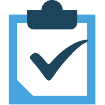}
    \fi
  }
  \fancyfoot[C]{}
}
\begin{document}
\maketitle

\ifdefined\hpcacameraready 
  \thispagestyle{camerareadyfirstpage}
  \pagestyle{empty}
\else
  \thispagestyle{plain}
  \pagestyle{plain}
\fi

\newcommand{\hpcaheight}{0mm}
\ifdefined\eaopen
\renewcommand{\hpcaheight}{12mm}
\fi


\begin{abstract}

  Retrieval-Augmented Generation (RAG) improves model output accuracy by leveraging external knowledge bases, serving as an effective solution to address hallucination issues and knowledge-update delays in Large Language Models (LLMs). However, the introduction of external knowledge bases presents RAG with challenges in long-context processing, significantly increasing memory consumption and inference latency. Existing research accelerates inference by precomputing Key and Value (KV)  of the knowledge base and loading them on-demand during inference. Based on the access frequency of different KV chunks within the external knowledge base, this paper proposes a hotness-aware RAG (HA-RAG) inference optimization system. First, leveraging the numerical distribution of KV chunks, we introduce a hotness-aware mixed-precision compressing and loading method to reduce disk I/O and memory access overhead. Second, we design a hotness-aware data placement strategy that prioritizes storing frequently accessed KV chunks in high-speed memory to improve data access efficiency. Experimental results demonstrate that, compared with TurboRAG, the proposed HA-RAG achieves an average speedup of 2.10$\times$ and maximum speedup of 10.49x in Time-To-First-Token (TTFT) with negligible accuracy loss.

\end{abstract}

\section{Introduction}\label{sec:intro}

With the advancement of artificial intelligence technologies, Large Language Models (LLMs) have demonstrated strong capabilities in tasks such as text generation, translation, and more,  leading to widespread applications across a variety of fields \cite{DBLP:journals/inffus/XiaoZLLLLH25, DBLP:conf/naacl/SonLKKMCPYB25, DBLP:conf/nips/HuangBZZZSLLZLF23, DBLP:journals/tacl/ZhangLDLMH24, DBLP:conf/naacl/ZhuLDXHKCL24}. However, LLMs have been observed to suffer from the hallucination problem in real-world scenarios, where the generated responses may contain information that deviates from factual truth \cite{DBLP:journals/tois/HuangYMZFWCPFQL25, DBLP:journals/tmlr/SadasivanKB0F25, DBLP:conf/ijcnlp/BangCLDSWLJYCDXF23, DBLP:journals/tacl/GuerreiroAWHBCM23}.
The issue primarily arises from two key factors: firstly, the inherent representational limitations of LLMs make it challenging to comprehensively and accurately capture complex semantic information; secondly, the time lag in updating training data hinders the timely integration of the latest knowledge into the models \cite{DBLP:journals/corr/abs-2310-05177}.

\begin{figure}[ht]
    \centering
    \includegraphics[width=0.99\linewidth]{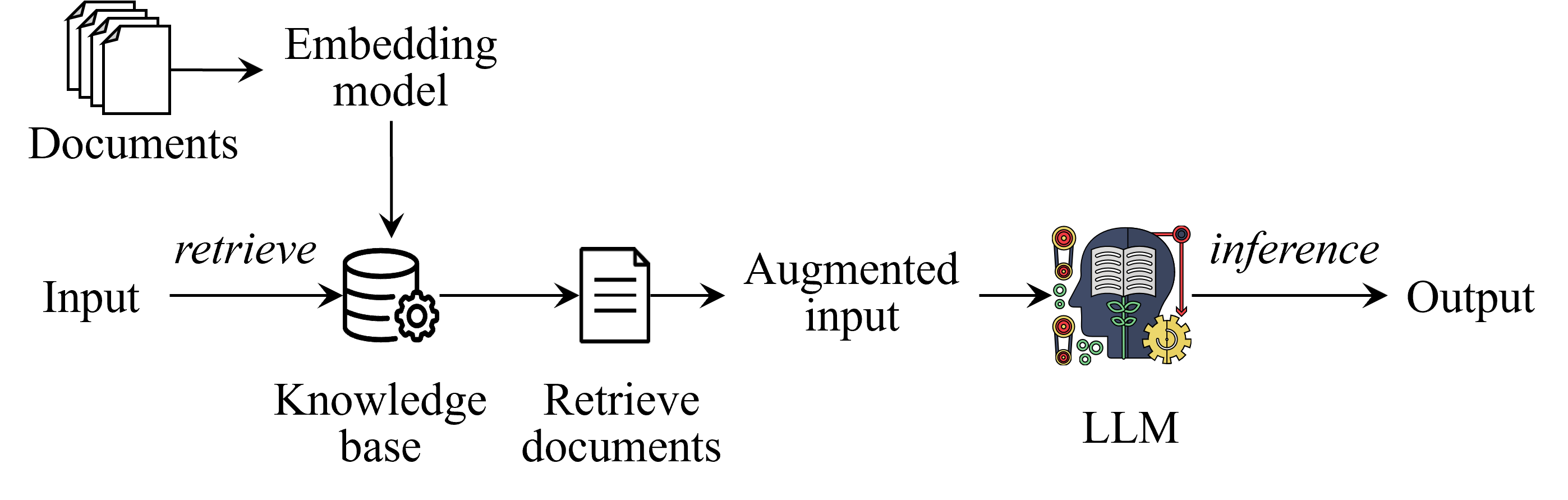}
    \caption{Workflow of RAG.}
    \label{RAG}
\end{figure}

Retrieval-Augmented Generation (RAG)~\cite{lewis2020retrieval} has emerged as an effective approach to mitigate this problem. By injecting task-specific contextual knowledge into the input, RAG significantly enhances the quality of generated text and the factual reliability of the output, thereby reducing hallucinations in LLMs \cite{DBLP:conf/kdd/FanDNWLYCL24}. As illustrated in Figure~\ref{RAG}, the typical workflow of RAG begins by embedding documents closely related to tasks into vector representations to form a knowledge base. Then, for each user input, a similarity measure such as cosine similarity is used to retrieve the most relevant documents from the knowledge base. These retrieved documents are appended to the input as additional context, resulting in an augmented input prompt, which is then fed into the LLM for inference. By incorporating context that is specific to downstream tasks, RAG provides more targeted and relevant information to the model. Even when the training data lacks examples related to a particular task, RAG enables the model to access key information via retrieval, thereby facilitating high-quality generation. 

However, RAG significantly extends the input length, which in turn leads to substantial computational overhead during the prefill stage of inference. Specifically, in the prefill phase, LLM must encode the entire input sequence and compute attention interactions between every pair of tokens to generate the first output token. This results in a computational complexity that grows quadratically with respect to the input length~\cite{dao2022flashattention}, severely limiting the generation efficiency in the RAG system. To address the challenge, TurboRAG~\cite{lu2024turborag} proposed to precompute the key-value (KV) of retrieved documents and store them on disk. During inference, these precomputed KV chunks are loaded on demand to avoid redundant computation, thereby accelerating the inference in the RAG system. 

While this approach significantly reduces the computational overhead in the prefill stage, RAG still faces two critical challenges. \textbf{First, it suffers from high data loading overhead}. Taking LLaMA2-7B~\cite{touvron2023llama} as an example, it consists of 32 layers, each with 32 attention heads and 128 dimensions per head. Assuming that the external documents are divided into 1024 chunks, with each chunk containing 512 tokens, and KV chunks are stored in BF16 format (2 bytes per value), the total disk space required for KV chunks amounts to approximately 128 GB. Clearly, transferring such a large amount of KV data from disk to GPU memory incurs significant overhead. We use MS MARCO~\cite{nguyen2016ms} as the retrieval corpus to generate 4,381 KV chunks and evaluate the RAG system on 256 queries randomly selected from the TriviaQA~\cite{joshi2017triviaqa} dataset. Results show that when generating a sequence of 16 tokens, the KV chunks loading time accounts for approximately 70\% of the total inference latency. \textbf{Second, current approaches have not sufficiently explored KV chunk scheduling optimization to improve memory access efficiency}. Since each KV chunk is typically large, transferring data between host and device memory incurs considerable latency. In practical RAG scenarios, the same KV chunk may be accessed multiple times. If every access requires reloading the chunk from disk and transferring it to GPU memory, it can result in substantial I/O and memory access overhead. Therefore, KV chunk scheduling is of great importance. Existing scheduling strategies in RAG primarily focus on pipeline parallelism between retrieval and generation~\cite{jin2024compute,yu2025ragdoll}, or on vector index-based dispatching mechanisms~\cite{jin2024ragcache}, aiming to balance retrieval latency and generation efficiency. However, scheduling strategies specifically tailored to KV chunk access patterns remain largely underexplored.

\begin{figure}[ht]
    \centering
    \includegraphics[width=0.9\linewidth]{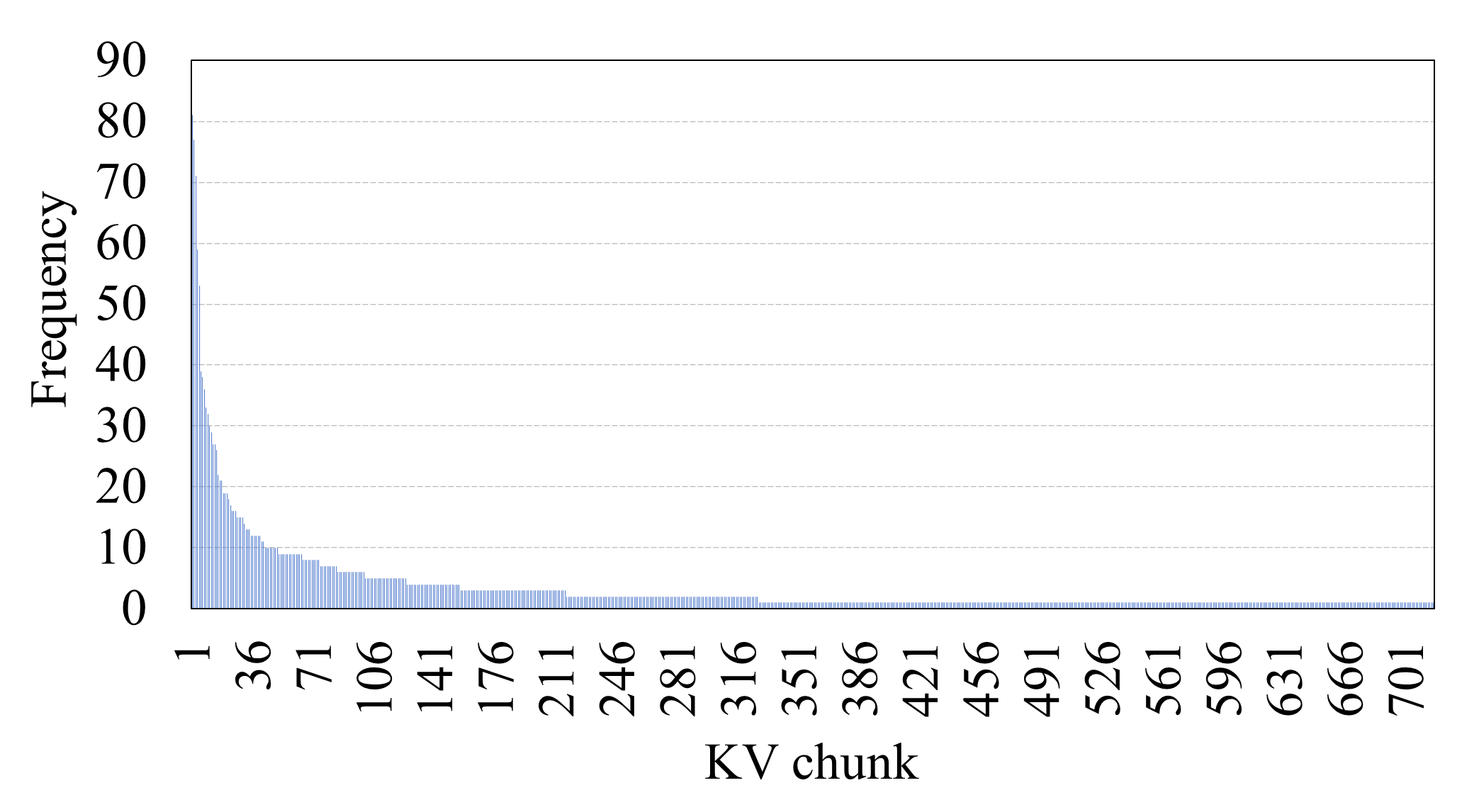}
    \caption{Access frequency of different KV chunks.}
    \label{fig:doc-frequency}
\end{figure}

In this work, we conduct a statistical analysis of access frequency across documents in the retrieval corpus. We find that only about 1\% of the documents are frequently retrieved (as shown in Figure~\ref{fig:doc-frequency}), revealing a highly skewed access frequency distribution. Motivated by this observation, we propose a hotness-aware RAG optimization framework (HA-RAG). By analyzing the value distribution characteristics of KV chunks and exploring various formats for low-precision data representation, we enable mixed-precision compressing and loading, thereby reducing the memory footprint and loading overhead of KV chunks. Furthermore, based on the access frequency of each KV chunk, we designed a data placement strategy that leverages heterogeneous hardware memory hierarchies to improve memory access efficiency during inference.

The main contributions of this paper are as follows:
\begin{itemize}
\item We propose a hotness-aware mixed-precision compression and loading strategy for KV chunks, which reduces data loading and memory access overhead.
\item We design a hotness-aware data placement strategy for KV chunks to enhance memory access efficiency during RAG inference.
\item Experimental evaluations on a RAG system built with the MS MARCO dataset and LLaMA2-7B model demonstrate that HA-RAG achieves an average performance improvement of 2.10$\times$ over TurboRAG in Time-To-First-Token (TTFT).
\end{itemize}

The remainder of this paper is organized as follows: Section 2 introduces HA-RAG, including the hotness-aware mixed-precision compression and loading of KV chunks, as well as the hotness-aware data placement optimization. Section 3 presents a comprehensive performance evaluation of HA-RAG. Section 4 reviews related work on RAG. Section 5 concludes the work.

\section{Methodology}\label{sec:intro}

\subsection{Overview}

\begin{figure*}[t]
    \centering
    \includegraphics[width=0.99\linewidth]{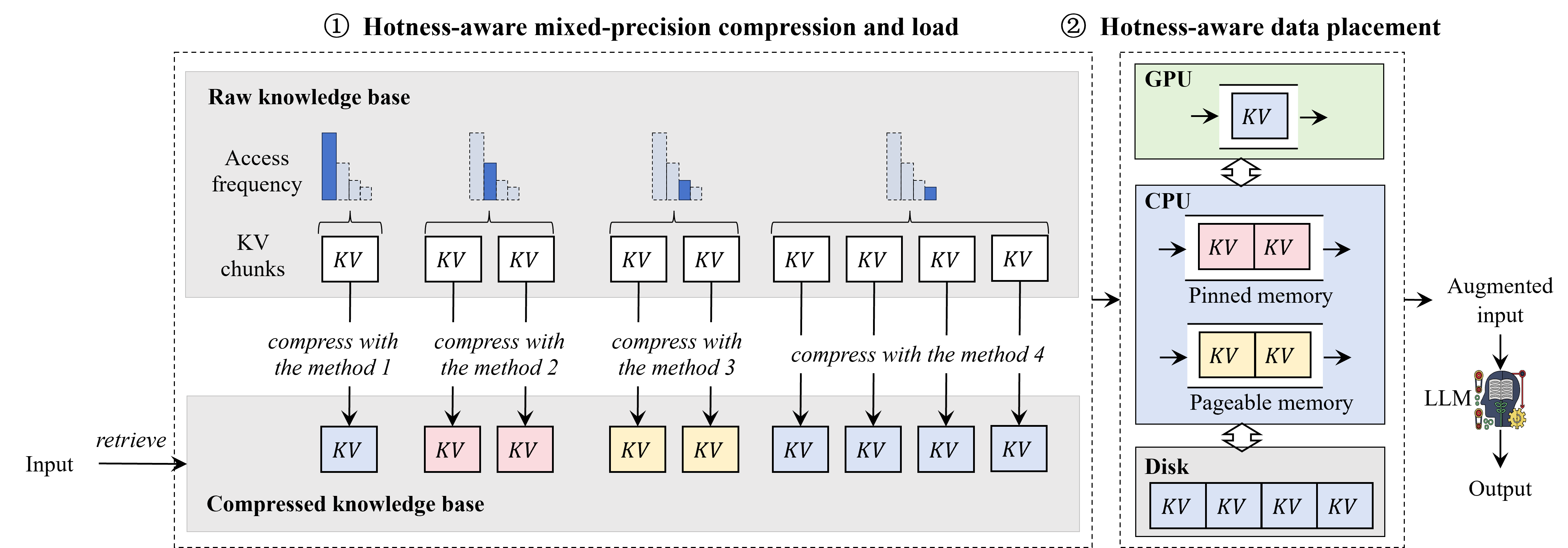} 
    \caption{Overall architecture of HA-RAG.}
    \label{fig:framework}
\end{figure*}

Figure~\ref{fig:framework} illustrates the overall architecture of the proposed HA-RAG. It consists of two main components: hotness-aware mixed-precision compression and loading, and hotness-aware data placement. For a given RAG system, we first obtain access frequency statistics for different KV chunks in the knowledge base by evaluating the system under real application scenarios. Based on the value distribution and access frequency of each KV chunk, the hotness-aware mixed-precision compression and loading module applies different compression methods to achieve mixed-precision compression, reducing the disk storage of these KV chunks. During inference, the compressed KV chunks are loaded into the system on demand, reducing loading time, memory usage, and memory access overhead associated with these KV chunks. Subsequently, the hotness-aware data placement module places frequently accessed KV chunks in fast-access memory (e.g., GPU or CPU pinned memory), while less frequently accessed chunks are stored in slower memory and loaded on demand (e.g., CPU pageable memory and disk), reducing the overall memory access overhead of the RAG system.

\subsection{Hotness-Aware Mixed-Precision Compression and Loading}
\label{sec:compression}

In this subsection, we first analyzed the numerical distribution characteristics of KV chunks in the knowledge base, including their value ranges and exponent distributions. We then apply several low-precision formats to compress KV chunks, including INT8, FP8 (both E4M3 and E5M2 formats), as well as a grouped shared exponent representation method (GSE) \cite{DBLP:journals/corr/abs-2411-04686}. Finally, we introduced a hotness-aware mechanism for KV chunks, where different low-precision formats are selected for different chunks based on their access frequencies.

\subsubsection{Numerical Characteristics Analysis}

\begin{figure}[t]
    \centering
    \includegraphics[width=0.99\linewidth]{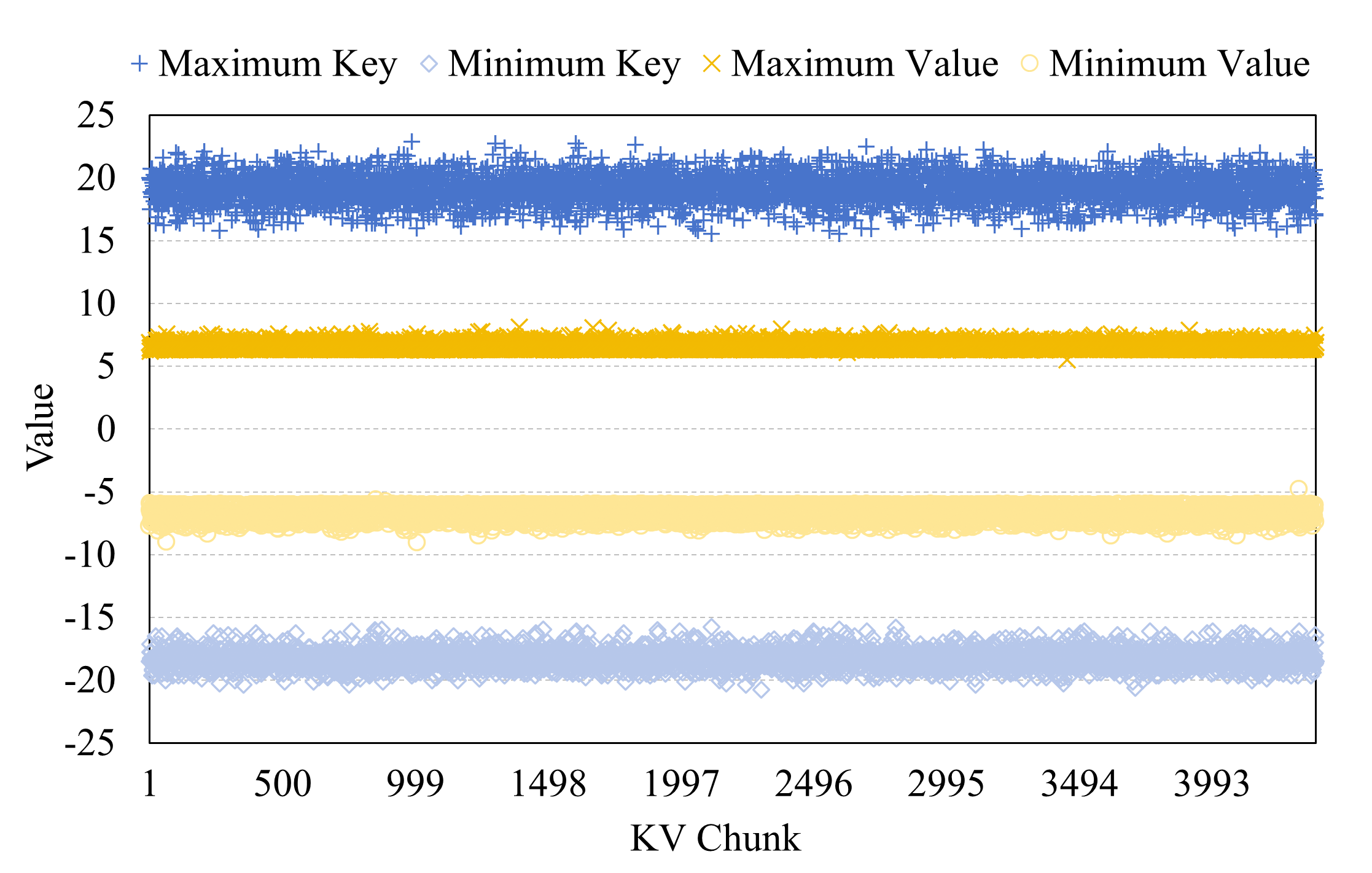}
    \caption{Value range distribution of Keys and Values in KV chunks.}
    \label{fig:KVChunk_range}
\end{figure}

We begin by performing a statistical analysis of the value ranges of Keys and Values from 4,381 KV chunks in the MS MARCO dataset. As shown in Figure~\ref{fig:KVChunk_range}, the values are relatively concentrated and exhibit narrow distributions. Specifically, all Key chunks primarily fall within the range of ($-25$, $25$), while Value chunks are distributed within ($-10$, $10$). Currently, KV chunks are typically stored in BF16 format, which consists of 1 sign bit, 8 exponent bits, and 7 fraction bits. The limited numerical range of KV chunks leads to significant bitwidth wastage when stored in BF16 format. This motivates the adoption of lower-precision data representations to compress and load KV chunks more efficiently.

\begin{figure}[t]
    \centering
    \includegraphics[width=0.99\linewidth]{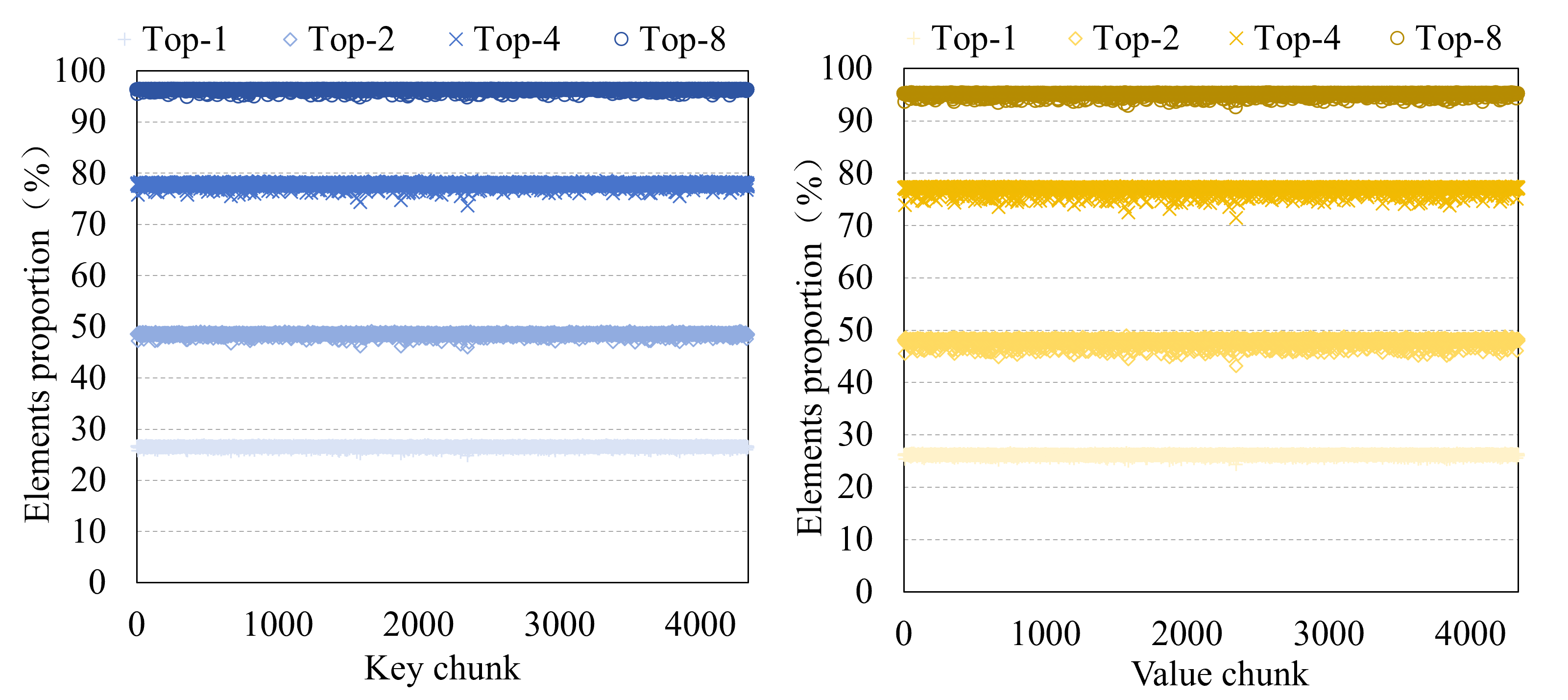}
    \caption{Exponent distribution of floating-point values in Key (left) and Value (right) chunks. Top-$k$ denotes the proportion of elements covered by the $k$ most frequent exponents.}
    \label{fig:KV-exponent-range}
\end{figure}

To further explore more efficient data representation methods for compressing KV chunks, we follow the approach proposed by Gao et al.\cite{DBLP:journals/corr/abs-2411-04686} and conduct a statistical analysis of the exponent distribution of floating-point values within KV chunks. The results are shown in Figure~\ref{fig:KV-exponent-range}.
As shown in the figure, the exponent distribution of floating-point elements in KV chunks is relatively concentrated. Specifically, in the Key chunks, $8$ exponents are sufficient to cover approximately 96\%-97\% of the elements. Similarly, in the Value chunks, the $8$ most frequent exponents can cover approximately 95\%-96\% of the elements. This indicates that conventional floating-point formats, which use fixed bit widths for both exponent and fraction, contain redundancy when applied to KV chunks, leading to inefficient use of limited bit-width resources. Motivated by this observation, we use the GSE format proposed in \cite{DBLP:journals/corr/abs-2411-04686} for compressing KV chunks.

\subsubsection{Low-Precision Data Representation}

\begin{figure}[t]
    \centering
    \includegraphics[width=0.95\linewidth]{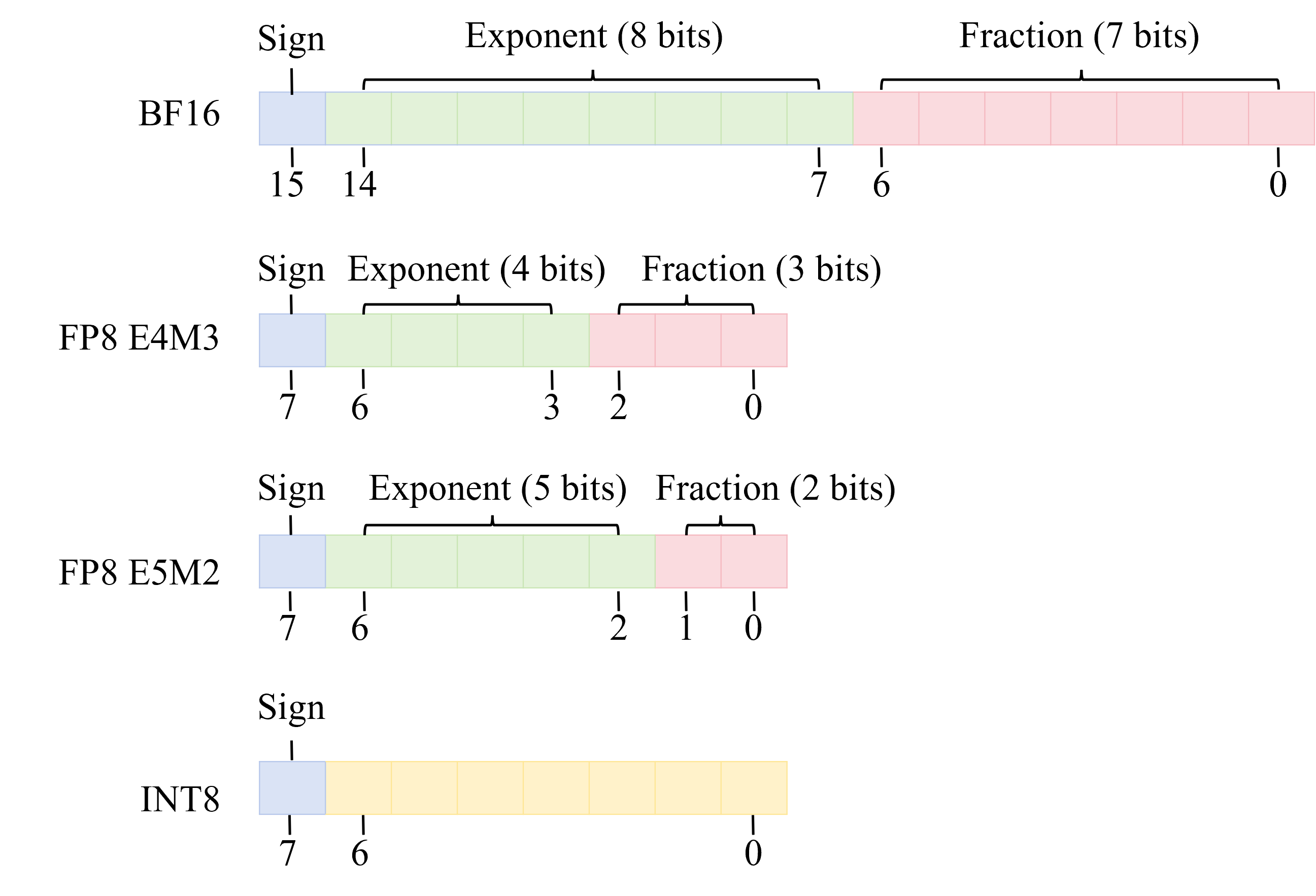}
    \caption{Comparison of different data representation formats.}
    \label{fig:fp8-format}
\end{figure}

Given that KV chunks are typically stored in the BF16 format, this work focuses on exploring $8$-bit low-precision data representations, including FP8, INT8, and GSE-8. FP8 \cite{DBLP:journals/corr/abs-2209-05433} is an $8$-bit floating-point format that has two common variants: E5M2 and E4M3, where ``E'' denotes the exponent bits and ``M'' denotes the fraction bits. As shown in Figure~\ref{fig:fp8-format}, E5M2 consists of $1$ sign bit, $5$ exponent bits, and $2$ fraction bits, and supports a representable range of approximately [$-57{,}334$, $57{,}334$]. In contrast, E4M3 consists of $1$ sign bit, $4$ exponent bits, and $3$ fraction bits, representing values in the range of approximately [$-448$, $448$]. E5M2 offers a wider dynamic range, while E4M3 provides higher numerical precision. INT8, commonly used to represent $8$-bit signed (or signed) integers, covers a range of [$-128$, $127$]. Given the relatively narrow value distribution observed in KV chunks, using such low-precision formats enables effective reduction in storage, loading, and memory access overhead, while maintaining acceptable accuracy and avoiding significant precision loss.

The GSE format is a novel floating-point representation designed for value-distribution-aware compression. It enables efficient encoding by sharing exponents across groups and couples storage with computation through segmented fraction storage. This format is primarily targeted at mixed-precision optimization in iterative algorithms for scientific computing. In the original design, authors firstly analyze the exponent frequency distribution among the nonzero elements of a sparse matrix and select the top-$k$ most frequent exponents to be stored in a shared exponent array. Subsequently, the nonzero values originally stored in FP64 format are converted into the proposed GSE representation. For elements whose exponent is not present in the shared exponent array, the nearest available shared exponent is used, and the fraction is adjusted accordingly via a denormalized transformation. To ensure that the GSE representation can be accurately reconstructed back to FP64, the index of the selected shared exponent must be recorded for each value. Authors encoded the exponent index into the column index field of each nonzero entry. During computation on the device side, the GSE representation is decoded into FP64 by referring to the shared exponent. Due to the long fraction length in FP64, the authors also introduced a fraction segmentation storage in the GSE format, which determines whether to load more tail segments based on the changes in residuals during the iteration process.

\begin{figure}[t]
    \centering
    \includegraphics[width=0.99\linewidth]{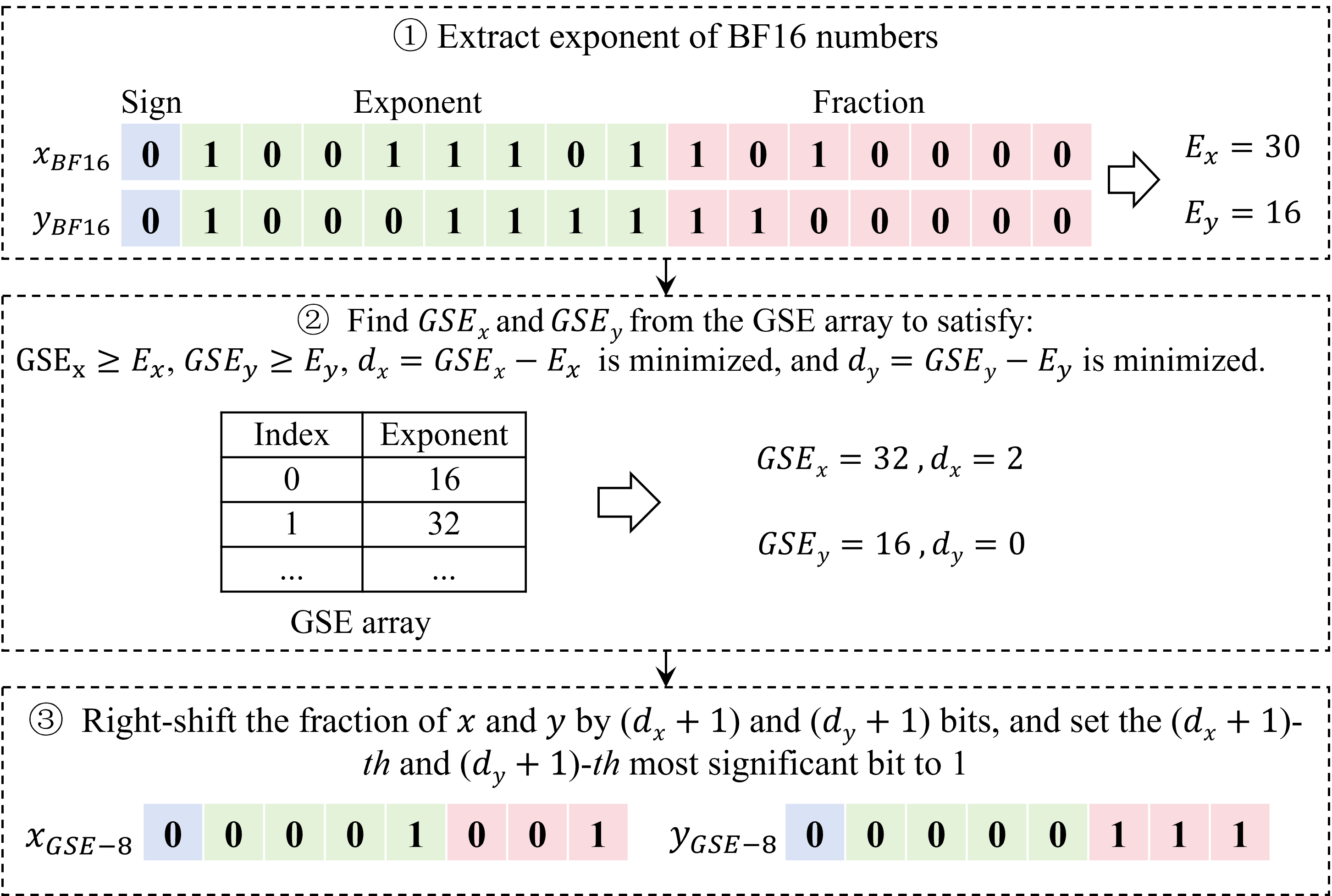}
    \caption{An example of converting BF16 numbers to GSE-8 format.}
    \label{fig:gse_encoding}
\end{figure}

Inspired by the concentrated exponent distribution observed in KV chunks, we adopt the GSE concept for compressing KV chunks. Different from the original GSE implementation, which typically uses $16\sim 64$ bits, we propose an $8$-bit variant referred to as GSE-8. For the selection of shared exponents, the original GSE chooses based on exponents frequency, whereas our method incorporates the numerical range of KV chunks to minimize the representation error. This design consideration is crucial, since using an unmatched exponent requires a denormalized fraction transformation, which introduces additional precision loss. As illustrated in Figure~\ref{fig:gse_encoding}, we take two BF16 numbers $x$ and $y$ as examples to introduce the conversion process into GSE-8. It includes the following three steps:

\begin{itemize}
    \item Extract the exponent fields of $x$ and $y$, denoted as $E_x$ and $E_y$, respectively.
    \item From the predefined GSE array (storing the shared exponents), find the smallest exponents greater than or equal to $E_x$ and $E_y$, denoted as $GSE_x$ and $GSE_y$, and compute the difference: $d_x = GSE_x-E_x$ and $d_y=GSE_y-E_y$.
    \item Right-shift the fraction of $x$ by $(d_x + 1)$ bits and that of $y$ by $(d_y + 1)$ bits. Then, set the $(d_x + 1)$-th and $(d_y + 1)$-th most significant bits of their fractions to $1$, obtaining the GSE-8 representation.
\end{itemize}

The decompression process is the reverse of compression. Based on the exponent index and the position of the first ``1" bit in the fraction field, the original exponent can be inferred. The fraction is then left-shifted accordingly to reconstruct an approximate version of the original value. During the conversion from BF16 to GSE-8, some precision loss occurs due to discarded low-order fraction bits during right-shifting. The number of discarded bits depends on the gap between the true exponent and the nearest larger shared exponent, making the selection of the shared exponent set critical for maintaining accuracy.

In traditional GSE formats, the total bit width is relatively large (typically $16\sim 64$ bits), and the index of the shared exponent is stored in the sparse matrix's column indices. As a result, the impact of precision loss caused by bit-shifting is minimal. However, in the GSE-$8$ format considered in this work, improper selection of shared exponents can lead to excessive shifting operations, which in turn cause significant precision degradation. To address this issue, we divide the distribution range of each KV chunk into sub-intervals using a fixed step size and assign one shared exponent to each sub-interval. For example, consider a KV chunk whose values lie within the range ($-10$, $10$). In GSE-$8$, we allocate $1$ bit for the sign, $3$ bits for the shared exponent index, and $4$ bits for the fraction. Due to the use of denormalized fractions, one bit must be reserved to explicitly represent the leading $1$, leaving $3$ effective bits for encoding. To avoid excessive precision loss caused by right-shifting, the shift distance must not exceed $3$ bits. This imposes a constraint that the difference between any two adjacent shared exponents should be no greater than $3$. Therefore, we divide the range ($-10$, $10$) into sub-intervals with a step size of $3$ and use the right endpoint of each interval as the shared exponent. The resulting shared exponent array is [$-7$, $-4$, $-1$, $2$, $5$, $8$, $10$], and the total number of exponents fits within the range representable by the $3$-bit shared exponent index.

\subsubsection{Hotness-Aware Mixed-Precision Compression and Loading}

As shown in Figure~\ref{fig:doc-frequency}, KV chunks exhibit different access frequencies. The core of the hotness-aware mixed-precision compression and loading strategy is to categorize KV chunks into ``hot'' and ``cold'' blocks based on their access frequency. ``Hot'' chunks refer to those that are frequently accessed, while ``cold'' chunks are infrequently accessed. Since the storage overheads of the various low-precision data formats considered are comparable, our mixed-precision design focuses on two aspects: accuracy loss and decompression overhead. For hot chunks, compression schemes with lower accuracy loss are preferred to ensure that the majority of inference tasks are not negatively affected, while still reducing loading and memory usage overhead. 
For cold chunks, we favor compression formats that offer shorter decompression time, in order to improve runtime efficiency without incurring significant online decompression overhead. This hotness-aware compression approach balances inference quality and I/O efficiency, leading to reduced memory and disk overhead with minimal loss in output quality.

Algorithm~\ref{alg:heataware} presents the hotness-aware mixed-precision compression algorithm. Among the four compression schemes, $S_1$ has the lowest precision loss and is applied to the most frequently accessed chunks, followed by $S_2$, while $S_4$ incurs the highest error but achieves the lowest decompression cost and is thus used for the coldest chunks. The thresholds $\tau_1$, $\tau_2$, and $\tau_3$ specify the proportions of KV chunks to be compressed with $S_1$, $S_2$, and $S_3$, respectively. The algorithm first sorts the KV chunks in descending order of access frequency based on $AF$ (line~\ref{alg:heataware:sort}). Then, based on the thresholds, the sorted chunks are divided into four groups (lines~\ref{alg:heataware:idx1}$\sim$\ref{alg:heataware:chunks4}), where $chunks_1$ represents the hottest chunks, followed by $chunks_2$ and $chunks_3$, and $chunks_4$ holds the least-accessed chunks. Finally, each group is compressed using its designated compression scheme (lines~\ref{alg:heataware:compress1}$\sim$\ref{alg:heataware:compress4}).

\begin{algorithm}
    \caption{Hotness-aware mix-precision compressing algorithm.}
    \label{alg:heataware}
    \KwIn{$KVChunks = [C_1^k, C_1^v, C_2^k, C_2^v, ..., C_n^k, C_n^v]$, access frequency of each KV Chunk $AF=[P_1, P_2, P_3, ..., P_{2n}]$, compression schemes: $S_1, S_2, S_3, S_4$, thresholds: $\tau_1, \tau_2, \tau_3$}
    \KwOut{Compressed KV chunks}  
    \tcp{\textbf{Step 1: Sort access frequency}}  
    $sortedChunks\gets Sort(KVChunks,AF,descending)$ \;\label{alg:heataware:sort}
    \tcp{\textbf{Step 2: Divide hot-cold KV chunks}}  
    $idx_1\gets \tau_1 \times 2n$ \;\label{alg:heataware:idx1}
    $idx_2\gets \tau_2 \times 2n + idx_1$ \;
    $idx_3\gets \tau_3 \times 2n + idx_2$ \;
    \tcp{the hottest chunks $chunks_1$}  
    $chunks_1\gets sortedChunks[0:idx_1]$ \; \label{alg:heataware:chunks1}
    \tcp{the second hottest chunks $chunks_2$}  
    $chunks_2\gets sortedChunks[idx_1:idx_2]$ \;
    \tcp{the third hottest chunks $chunks_3$}  
    $chunks_3\gets sortedChunks[idx_2:idx_3]$ \;
    \tcp{the coldest chunks $chunks_4$}  
    $chunks_4\gets sortedChunks[idx_3:2n]$ \;\label{alg:heataware:chunks4}
    \tcp{\textbf{Step 3: assign compression scheme for each chunk}}
    $compressChunks_1\gets Compression(S_1, chunks_1)$ \;\label{alg:heataware:compress1}
    $compressChunks_2\gets Compression(S_2, chunks_2)$ \;\label{alg:heataware:compress2}
    $compressChunks_3\gets Compression(S_3, chunks_3)$ \;\label{alg:heataware:compress3}
    $compressChunks_4\gets Compression(S_4, chunks_4)$ \;\label{alg:heataware:compress4}
\end{algorithm}

\subsection{Hotness-Aware Data Placement Optimization}

\begin{figure}[t]
    \centering
    \includegraphics[width=0.95\linewidth]{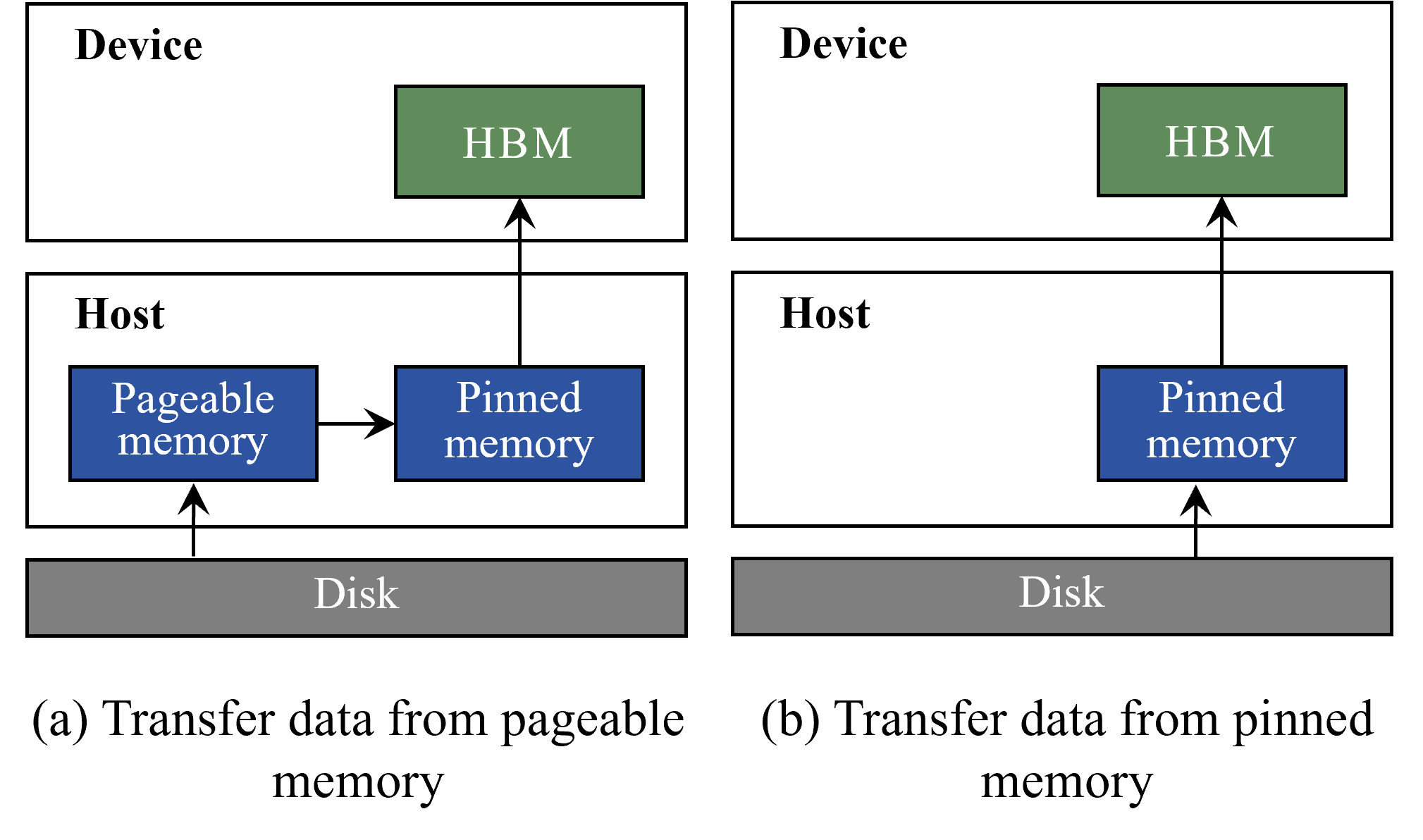}  
    \caption{Two transmission paths for data from disk to device memory in heterogeneous systems.}
    \label{fig:memory_path}
\end{figure}

During LLMs inference, the core computations are performed on GPU devices. As a result, KV chunks in RAG must be loaded progressively from disk into GPU memory before they can be used for computation. As shown in Figure~\ref{fig:memory_path}, this process involves multiple stages of memory transfer. 

By default, the operating system (OS) allocates memory on the CPU as pageable memory, where data is stored in virtual memory pages. When a program accesses a page that is not currently in RAM, a page fault is triggered, prompting the OS to swap the required page from disk into RAM. In contrast, pinned (or page-locked) memory cannot be swapped out to disk and offers faster and more consistent access latency. Data transfers from CPU memory to GPU memory typically fall into two scenarios: if the data resides in pinned memory, it can be directly transferred to the GPU with low latency. However, if the data is in pageable memory, it must first be copied to pinned memory before being sent to the GPU. Moreover, if the required page is not in RAM, it must first be fetched from disk into RAM, introducing significant transfer overhead.

The hotness-aware mixed-precision compression and loading scheme presented in the previous subsection effectively reduces memory footprint and disk I/O overhead by compressing KV chunks using low-precision formats. Building on the data transfer analysis from Figure \ref{fig:memory_path}, this subsection introduces a hotness-aware data placement strategy to further optimize overall memory access efficiency. The core idea is to allocate KV chunks across different memory hierarchies based on their access frequency and the characteristics of the memory hierarchy. Frequently accessed chunks (``hot" blocks) are prioritized for placement in high-speed storage, such as GPU memory and CPU pinned memory. Infrequently accessed chunks (``cold" blocks), by contrast, are stored in slower storage, such as CPU pageable memory or disk. By avoiding frequent evictions of hot chunks to slower storage, this strategy improves memory efficiency and system performance.

Algorithm~\ref{alg:data_placement} outlines the proposed hotness-aware data placement algorithm. The queues $queueGPU$, $queuePIN$, and $queuePAGE$ represent the sets of KV chunks currently residing in GPU memory, CPU pinned memory, and CPU pageable memory, respectively. The thresholds $\tau_{GPU}$, $\tau_{PIN}$, and $\tau_{PAGE}$ define the maximum number of KV chunks. First, as shown in lines~\ref{alg:data_placement:idx}$\sim$\ref{alg:data_placement:disk}, the sorted KV chunks (ranked by access frequency from Algorithm~\ref{alg:heataware}) are partitioned into four disjoint ranges, whose indices are stored in $GPU\_LIST$, $PIN\_LIST$, $PAGE\_LIST$, and $DISK\_LIST$. Chunks in $GPU\_LIST$ are the hottest, while those in $DISK\_LIST$ are the coldest. Next, for a given input query, the required KV chunks are retrieved (line~\ref{alg:data_placement:retrieve}). For each chunk $C_i$ (line~\ref{alg:data_placement:loop}), the system first checks if it is already residing in GPU memory and accesses it directly if so (line~\ref{alg:data_placement:gpu_hit}). If not, it checks whether $C_i$ is in CPU pinned memory (line~\ref{alg:data_placement:pin_hit}); if so, it loads the chunk from pinned memory into GPU memory (line~\ref{alg:data_placement:promote_gpu_from_pin}). Moreover, if its index is in $GPU\_LIST$, the algorithm removes the least recently used chunk via an LRU policy (if $queueGPU$ is full) and places the chunk into GPU memory. If the chunk is instead found in pageable memory (line~\ref{alg:data_placement:page_hit}), it is loaded for computation, and may also be promoted to GPU or pinned memory (lines~\ref{alg:data_placement:promote_gpu_from_page}$\sim$\ref{alg:data_placement:promote_pin_from_page}) depending on its access frequency. Finally, if the chunk is not found in any memory, it is loaded from disk (line~\ref{alg:data_placement:disk_load}), and then placed into the appropriate queue ($queueGPU$, $queuePIN$, or $queuePAGE$) according to its index in $GPU\_LIST$, $PIN\_LIST$, or $PAGE\_LIST$ (lines~\ref{alg:data_placement:promote_gpu_from_disk}$\sim$\ref{alg:data_placement:promote_page_from_disk}).

\begin{algorithm}
    \caption{Hotness-aware data placement algorithm.}
    \label{alg:data_placement}
    \KwIn{input query $query$, queues storing KV chunks: $queueGPU$, $queuePIN$, and $queuePAGE$, threshold: $\tau_{GPU}$, $\tau_{PAGE}$, $\tau_{PIN}$}
    \KwOut{Inference results with RAG.}   

    \tcp{\textbf{Step 1: divide sorted chunks}}
    $idx_1\gets \tau_{GPU} \times 2n$, $idx_2\gets \tau_{PAGE} \times 2n + idx_1$, $idx_3\gets \tau_{PIN} \times 2n + idx_2$ \; \label{alg:data_placement:idx}
    $GPU\_LIST\gets [0, 1, 2,...,idx_1-1]$ \; \label{alg:data_placement:gpu}
    $PIN\_LIST\gets [idx_1, idx_1+1, ..., idx_2-1]$ \; \label{alg:data_placement:pin}
    $PAGE\_LIST\gets [idx_2, idx_2+1, ..., idx_3-1]$ \; \label{alg:data_placement:page}
    $DISK\_LIST\gets [idx_3, idx_3+1, ..., 2n]$ \; \label{alg:data_placement:disk}

    \tcp{\textbf{Step 2: place KV chunks in appropriate memory}}
    $ chunks\gets retrieve\_documents(query)$ \; \label{alg:data_placement:retrieve}
    \ForEach{$C_i$ in $chunks$}{ \label{alg:data_placement:loop}
        \If{$C_i$ in $queueGPU$}{
    	    $queueGPU.get(C_i)$ \; \label{alg:data_placement:gpu_hit}
        }
        \ElseIf{$C_i$ in $queuePIN$}{
            $queuePIN.get(C_i)$ \; \label{alg:data_placement:pin_hit}
            \If{index of $C_i$ in $GPU\_LIST$}{
                put $C_i$ in $queueGPU$  with LRU \; \label{alg:data_placement:promote_gpu_from_pin}
            }
        }
        \ElseIf{$C_i$ in $queuePAGE$}{
            $queuePAGE.get(C_i)$ \; \label{alg:data_placement:page_hit}
            \If{index of $C_i$ in $GPU\_LIST$}{
                put $C_i$ in $queueGPU$ with LRU \; \label{alg:data_placement:promote_gpu_from_page}
            }
            \If{index of $C_i$ in $PIN\_LIST$}{
                $queuePIN.put(C_i)$ \; \label{alg:data_placement:promote_pin_from_page}
            }
        }
        \Else{		
            load $C_i$ from Disk \; \label{alg:data_placement:disk_load}
            \If{index of $C_i$ in $GPU\_LIST$}{
                put $C_i$ in $queueGPU$  with LRU \; \label{alg:data_placement:promote_gpu_from_disk}
            }
            \If{index of $C_i$ in $PIN\_LIST$}{
                $queuePIN.put(C_i)$ \; \label{alg:data_placement:promote_pin_from_disk}
            }
            \If{index of $C_i$ in $PAGE\_LIST$}{
                put $C_i$ in $queuePAGE$  with LRU \; \label{alg:data_placement:promote_page_from_disk}
            }
        }
    }
\end{algorithm}

\section{Experiments}

\subsection{Experimental Setup}

\textbf{Software and Hardware:} All experiments are conducted on a machine equipped with an NVIDIA A100 40GB GPU. Table~\ref{hardware} provides a detailed overview of the hardware and software configurations of the system.

\begin{scriptsize}
\begin{table}[h!]
\centering
\caption{System hardware and software configuration.}
\label{hardware}
\newcolumntype{Z}[1]{>{\centering\arraybackslash}p{#1}} 
\begin{tabular}{|Z{10mm}|Z{15mm}|p{40mm}|}
    \hline
    \textbf{Type} & \textbf{Component} & \textbf{Configuration} \\
    \hline
    \hline
    \multirow{5}{*}{Hardware} 
        & \multirow{3}{*}{Host} 
        & CPU: Intel Xeon Platinum 8336C \\
        \cline{3-3}
        & & Memory: 516 GB \\
        \cline{3-3}
        & & PCIe: PCIe 4.0 x16/x8 \\
        \cline{2-3}
        & \multirow{2}{*}{Device} 
        & GPU: NVIDIA A100 \\
        \cline{3-3}
        & & Memory: 40 GB \\
    \hline
    \multirow{2}{*}{Software} 
        & OS & Ubuntu 22.04 \\
        \cline{2-3}
        & Library & PyTorch: 2.0.1, CUDA: 11.5 \\
    \hline
\end{tabular}
\end{table}
\end{scriptsize}

\textbf{Model and Dataset:} The experiments use LLaMA-2-7B as the LLM and facebook/contriever \cite{DBLP:journals/tmlr/IzacardCHRBJG22} as the embedding model to convert retrieved documents into a vectorized knowledge base. We use 1,000 articles from MS MARCO as the retrieval corpus and TriviaQA as the test set. Documents in MS MARCO are segmented into chunks of 512 tokens per document.

\textbf{Performance Baseline}: We compare our method with TurboRAG \cite{lu2024turborag}, which precomputes KV chunks of the knowledge base, stores them on disk, and loads them on-demand during inference. In our experiments, TurboRAG stores KV chunks in BF16 format.

\textbf{Evaluation Metrics:}
Time-to-First-Token (TTFT) measures the latency between receiving an input and generating the first output token. Since RAG expands the input context with external documents, the prefill stage becomes slower. Our method primarily optimizes KV chunk loading overhead, making TTFT a key metric to evaluate our improvements. The second metric is ROUGE-1 F1 \cite{lin2004rouge}.
ROUGE (Recall-Oriented Understudy for Gisting Evaluation) assesses the quality of LLM-generated text. ROUGE-1 measures unigram overlap between generated and reference texts, with three key sub-metrics: ROUGE-1 Recall, ROUGE-1 Precision, and ROUGE-1 F1 (harmonic mean of recall and precision). We use ROUGE-1 F1 to evaluate the impact of different methods on the accuracy of RAG systems.

\subsection{Parameter Configuration}

\subsubsection{Comparison of Efficiency and Precision Loss among Compression Methods}

We apply various compression methods to KV chunks originally represented in BF16 format and evaluate their compression ratios and decompression times on a dataset of size $1024$, as summarized in Table~\ref{tab:compression_comparison}. Additionally, the GSE-$8$ format supports multiple encoding schemes based on the bit widths assigned to the exponent and fraction, such as $1+2+5$, $1+3+4$, and $1+4+3$, where the first digit represents the sign bit, the second represents the exponent width, and the third represents the fraction width. We conduct experiments on these three variants and find that the $1+4+3$ configuration yields the smallest error. Therefore, this configuration is adopted in our following experiments. As shown in Table~\ref{tab:compression_comparison}, GSE-$8$ achieves the lowest decompression latency, followed by INT8. In contrast, both FP8 formats exhibit higher decompression overhead. Since all methods use $8$-bit representations, the resulting compression ratios are approximately the same.

\begin{table}
  \centering
  \caption{Comparison of decompression time and compression ratio for different compression methods.}
  \label{tab:compression_comparison}
  \begin{tabular}{|>{\centering\arraybackslash}p{2cm}|>{\centering\arraybackslash}p{2cm}|>{\centering\arraybackslash}p{2cm}|}
    \hline
    \textbf{Compression method} & \textbf{Decompression time (s)} & \textbf{Compression ratio} \\
    \hline
    \hline
    FP8 E4M3 & 284.86 & 1.9996 \\
    \hline
    FP8 E5M2 & 254.72 & 1.9996 \\
    \hline
    INT8     & 20.23  & 1.9988 \\
    \hline
    GSE-8    & 17.38  & 1.9996 \\
    \hline
  \end{tabular}
\end{table}

\begin{figure}[t]
    \centering
    \includegraphics[width=0.98\linewidth]{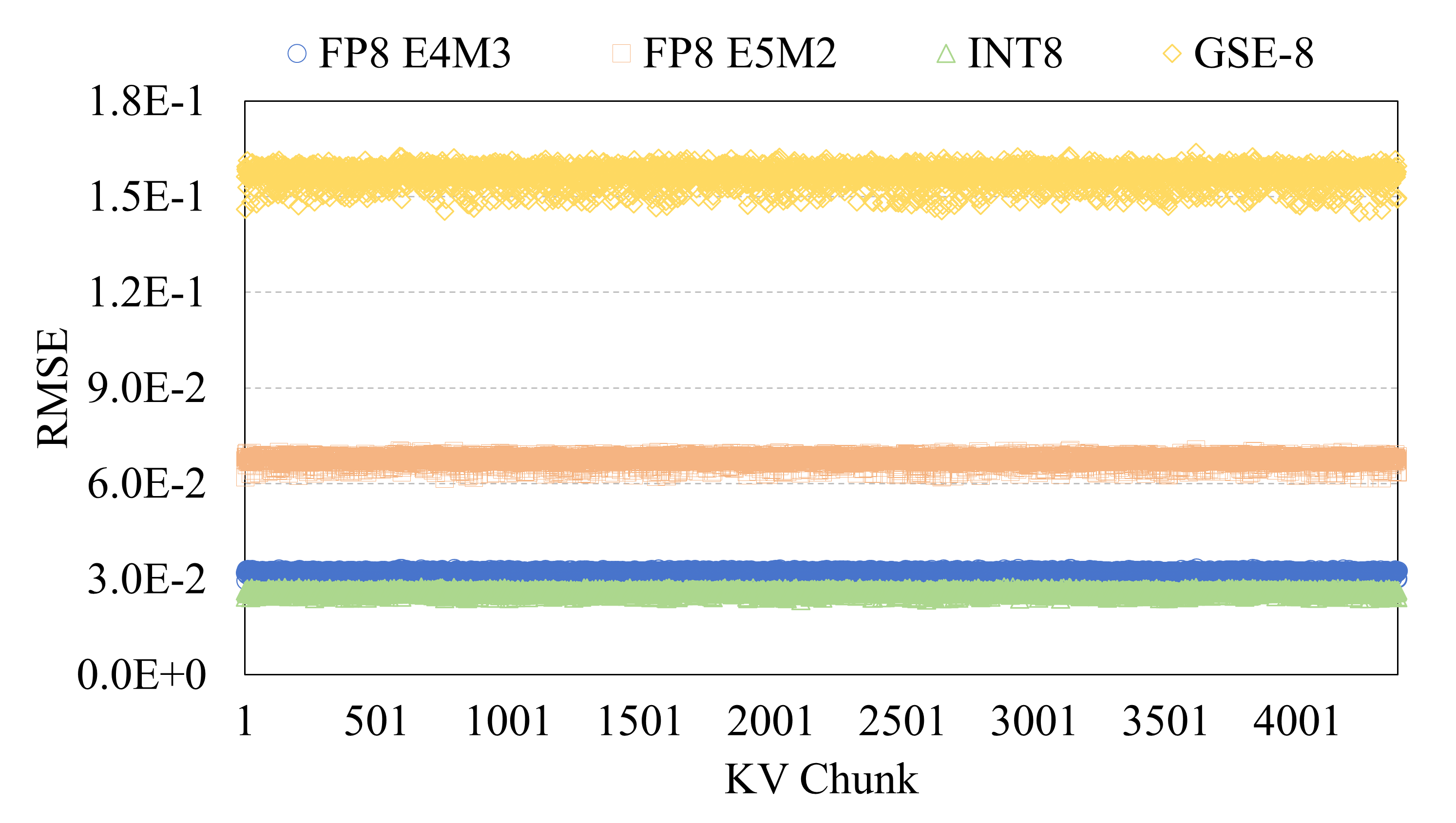}
    \caption{Root mean square error of different compression methods.}
    \label{fig:rmse_comparison}
\end{figure}

Furthermore, we evaluate the accuracy loss introduced by different compression methods using the Root Mean Square Error (RMSE) between the decompressed and original data, as defined in Equation~(\ref{eq:rmse}). Here, $x_i$ denotes the decompressed value, $\hat{x}_i$ is the original value before compression, and $N$ represents the number of elements in each KV chunk. Figure~\ref{fig:rmse_comparison} illustrates the RMSE results across various compression methods. It can be observed that INT8 yields the smallest accuracy loss, followed by E4M3. Due to its smaller fraction width, E5M2 exhibits a larger error than E4M3. GSE-$8$ results in the highest accuracy loss.

\begin{equation}\label{eq:rmse}
RMSE = \sqrt{\frac{1}{N} \sum_{i=1}^{N} (x_i - \hat{x}_i)^2}
\end{equation}

\begin{figure}[t]
    \centering
    \includegraphics[width=0.98\linewidth]{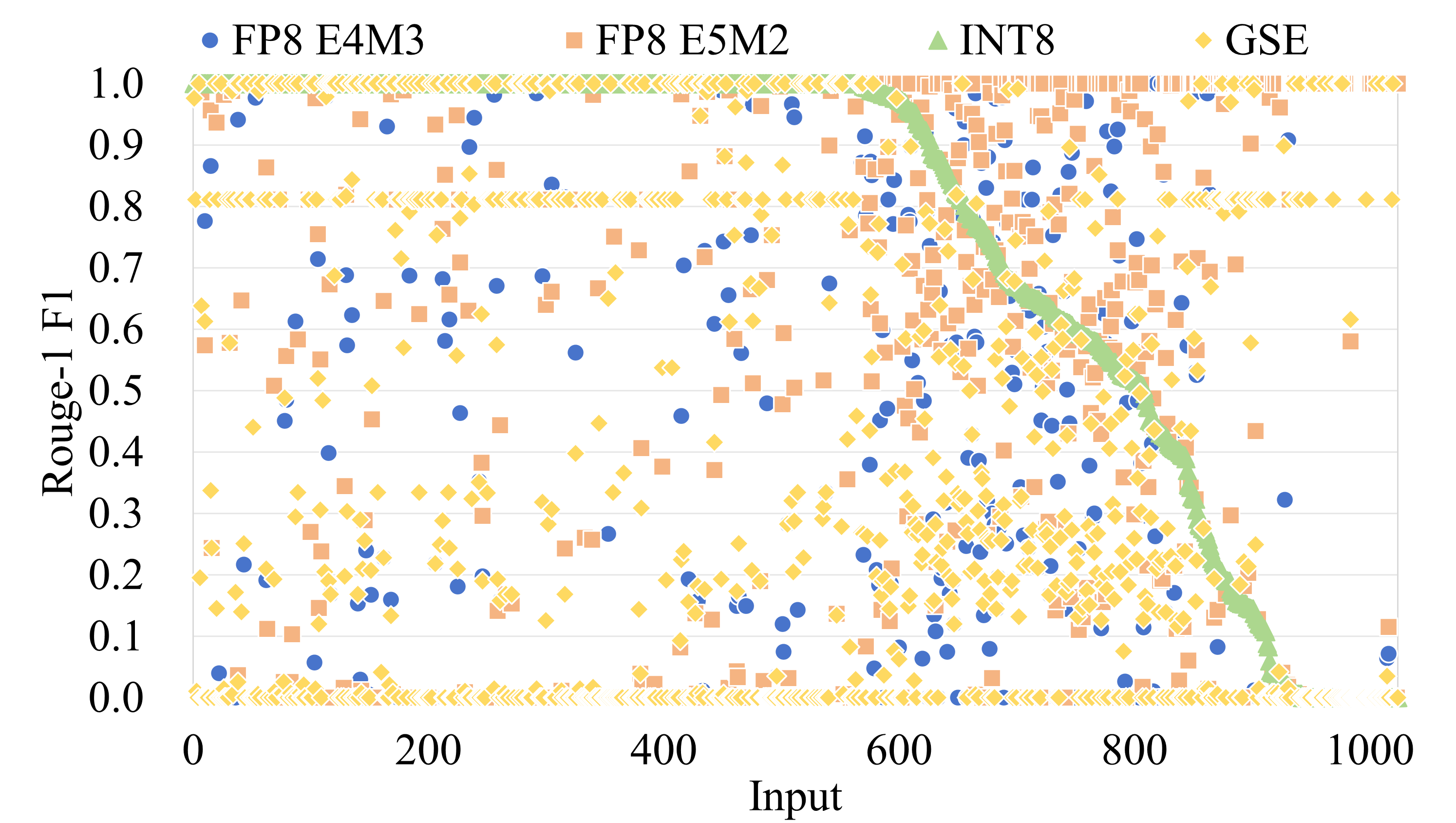}
    \caption{The impact of each compression method on the inference accuracy of the RAG system.}
    \label{fig:ROUGE_comparison}
\end{figure}

\begin{figure}[t]
    \centering
    \subfloat[E4M3]{
        \includegraphics[width=0.49\linewidth]{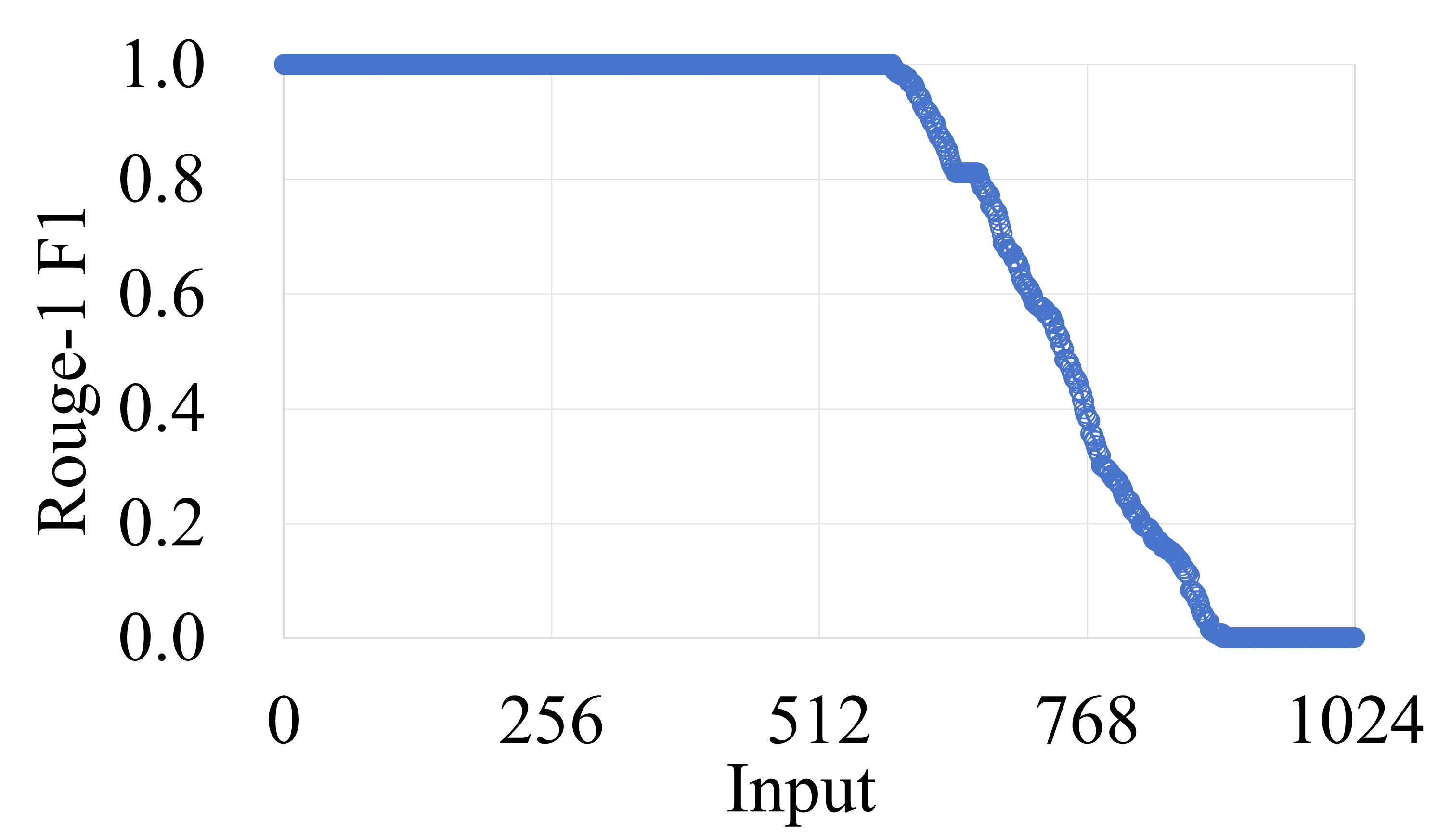}
        \label{fig:rouge_E4M3}
    }
    \subfloat[E5M2]{
        \includegraphics[width=0.49\linewidth]{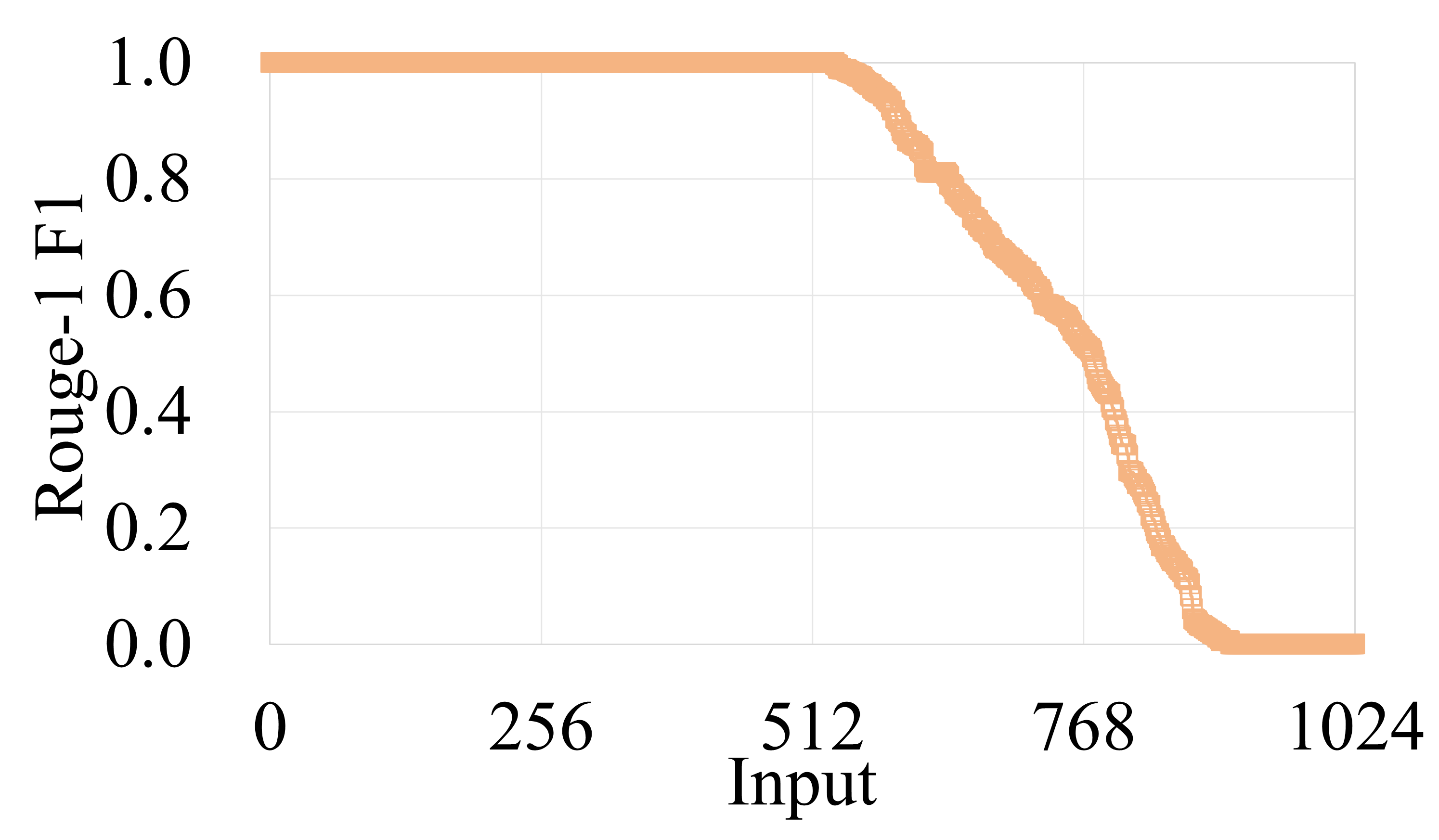}
        \label{fig:rouge_E5M2}
    }
    \\
    \subfloat[INT8]{
        \includegraphics[width=0.49\linewidth]{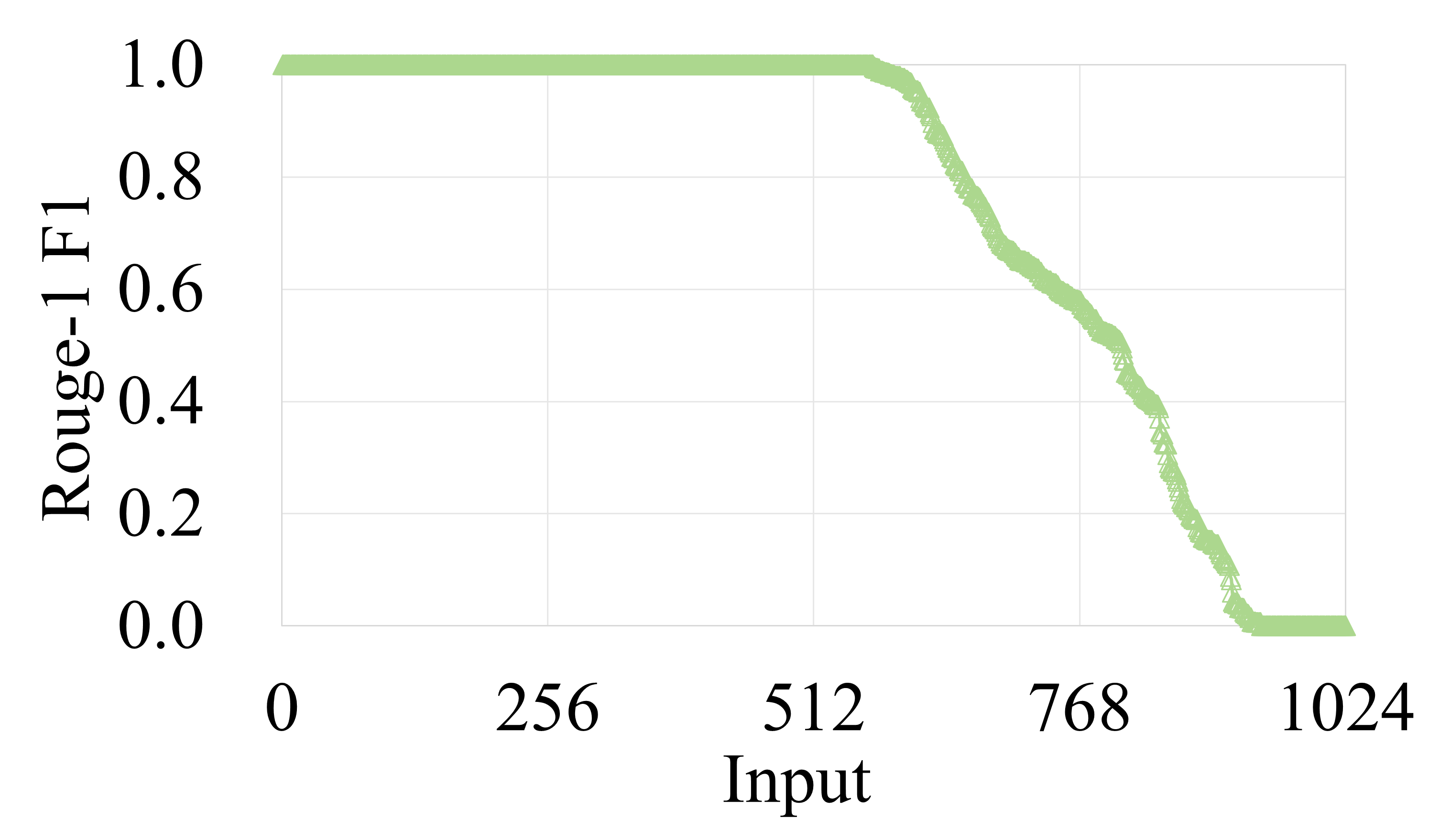}
        \label{fig:rouge_INT8}
    }
    \subfloat[GSE-8]{
        \includegraphics[width=0.49\linewidth]{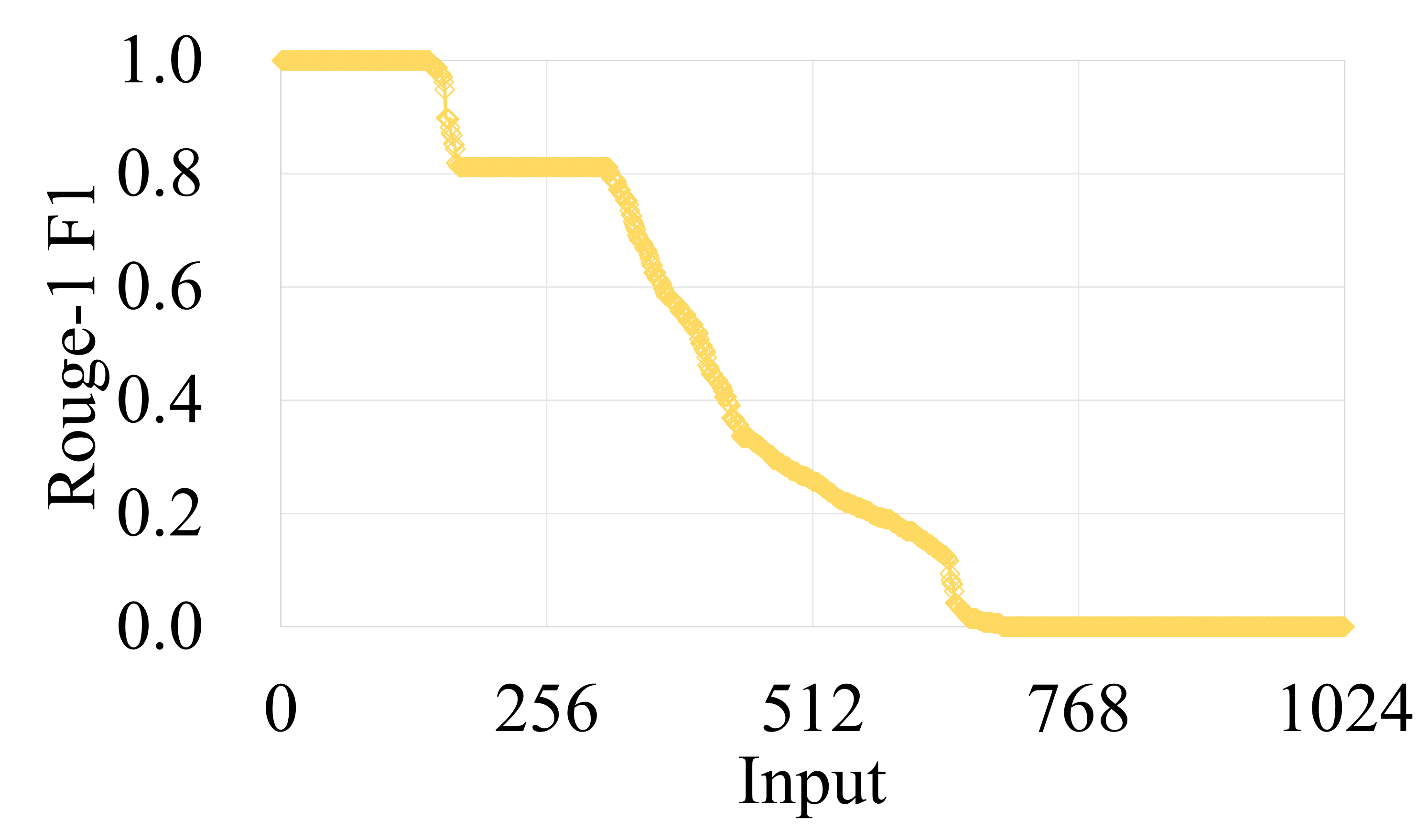}
        \label{fig:rouge_GSE}
    }
    \caption{Comparison of the inference accuracy of the RAG system under different compression methods, ranked by ROUGE-1 F1 score in descending order.}
    \label{fig:single_4}
\end{figure}

To further investigate the impact of different compression methods on the inference quality of RAG systems, we evaluate the accuracy of the RAG when using each compression method individually on 1024 test queries. Figure~\ref{fig:ROUGE_comparison} and Figure~\ref{fig:single_4} illustrate the ROUGE-1 F1 scores of each method relative to the TurboRAG. In Figure~\ref{fig:ROUGE_comparison}, the results are sorted in descending order based on the ROUGE-1 F1 scores of the FP8 format. In Figure~\ref{fig:single_4}, results are sorted in descending order based on respective ROUGE-1 F1 scores. 

We can observe from Figure~\ref{fig:single_4} that the overall trend of ROUGE-1 F1 scores aligns with the RMSE patterns observed in Figure~\ref{fig:rmse_comparison}. Specifically, INT8 achieves the smallest accuracy degradation, with over 80\% of the KV chunks yielding a ROUGE-1 F1 score above 0.5. E4M3 and E5M2 follow, while GSE-8 shows the lowest performance, with about 40\% of KV chunks achieving a ROUGE-1 F1 score above 0.5. Nevertheless, one interesting observation from the Figure~\ref{fig:ROUGE_comparison} is that the impact of compression methods on ROUGE-1 F1 is not always consistent with the general trend of GSE-8 $>$ E5M2 $>$ E4M3 $>$ INT8. For certain KV chunks, GSE-8 surprisingly introduces less accuracy degradation than the other compression methods. 

Based on the above experimental observations, we conclude that the four compression methods each have their own advantages in terms of efficiency and accuracy. Therefore, we use a combination of the four methods to compress KV chunks. Specifically, we use INT8 to compress the most frequently accessed KV chunks, which minimizes accuracy while achieving high decompression efficiency. For the next most frequently accessed KV chunks, we use E4M3 and then E5M2 for compression. For the least frequently accessed KV chunks, we use GSE-8 for compression.

\subsubsection{Threshold Parameter Selection}

Based on the observations in the previous subsection regarding the efficiency and accuracy of the four compression methods, we can interpret the threshold parameters $\tau_1$, $\tau_2$, and $\tau_3$ in Algorithm~\ref{alg:heataware} as corresponding to KV chunks compressed using INT8, E4M3, and E5M2, respectively. In this experiment, we set the following four groups of thresholds to investigate how different parameter settings affect final performance and accuracy.
\begin{itemize}
    \item $Param1 = \{\tau_1:10\%,\tau_2:5\%,~\tau_3:5\%\}$
    \item $Param2 = \{\tau_1:10\%,\tau_2:10\%,~\tau_3:5\%\}$
    \item $Param3 = \{\tau_1:10\%,\tau_2:10\%,~\tau_3:10\%\}$
    \item $Param4 = \{\tau_1:15\%,\tau_2:10\%,~\tau_3:10\%\}$
\end{itemize}

\begin{figure}[t]
    \centering
    \includegraphics[width=0.9\linewidth]{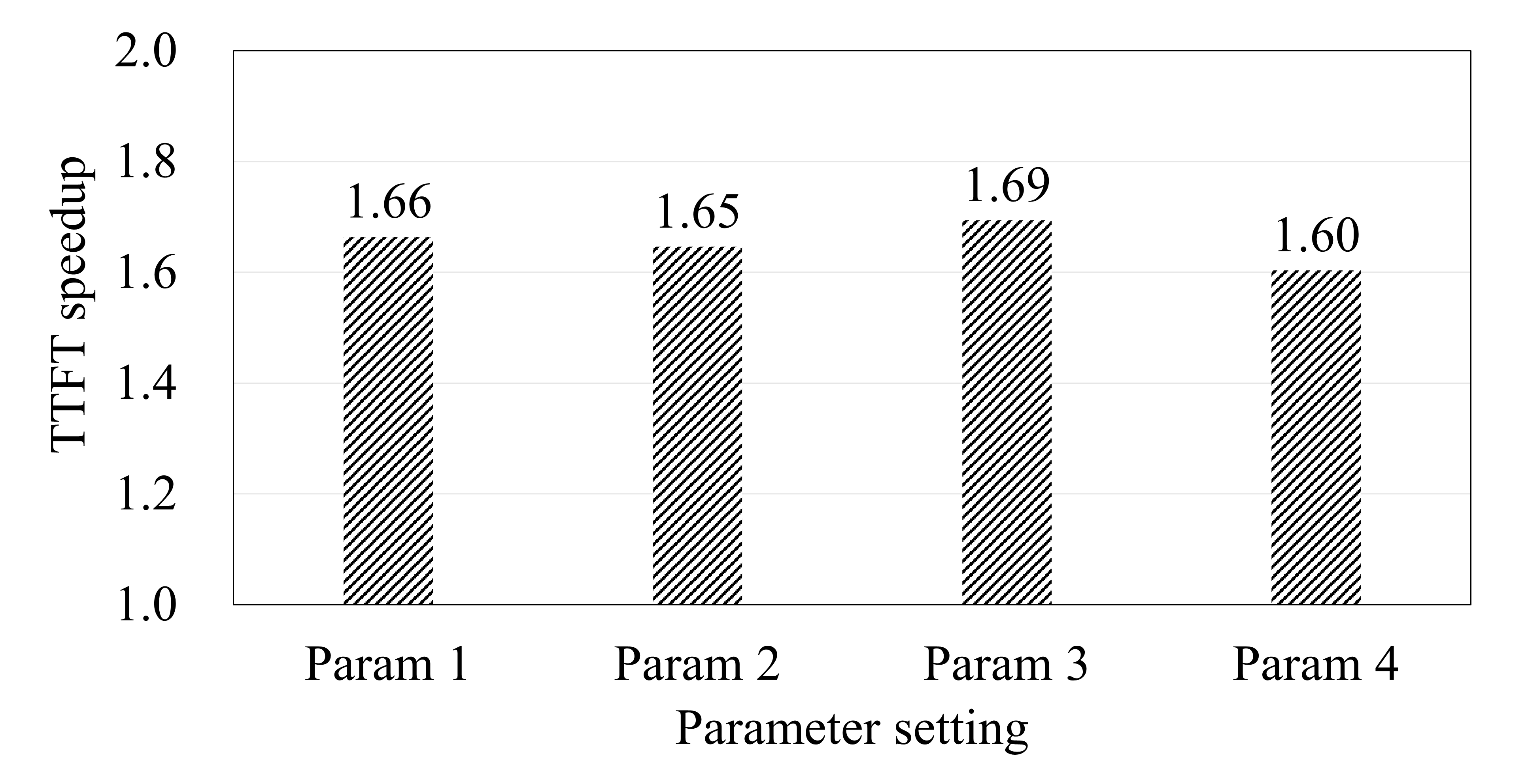}
    \caption{Performance comparison of $\tau_1$, $\tau_2$, and $\tau_3$ under four parameter configurations.}
    \label{config_TTFT}
\end{figure}

Additionally, in this experiment, the three thresholds in Algorithm~\ref{alg:data_placement} are fixed as $\tau_{GPU}=5\%$, $\tau_{PIN}=5\%$, and $\tau_{PAGE}=10\%$. Figure~\ref{config_TTFT} and Figure~\ref{fig:config_ROUGE} illustrate the impact of the four parameter configurations on the efficiency and accuracy of RAG inference (size of test set is 1024). It can be observed that the third parameter configuration achieves the highest inference efficiency. As shown in Figure~\ref{fig:config_ROUGE}, the third parameter configuration also results in comparable inference accuracy compared with others.

In this experiment, we fix the thresholds in Algorithm~\ref{alg:heataware} to the above first parameter configuration and evaluate the impact of the three thresholds in Algorithm~\ref{alg:data_placement}, $\tau_{GPU}$, $\tau_{PIN}$, and $\tau_{PAGE}$, on the inference efficiency of the RAG system. We consider the following two groups of parameter configurations. 

\begin{itemize}
    \item $Param1 = \{\tau_{GPU}:5\%,\tau_{PIN}:5\%,~\tau_{PAGE}:10\%\}$
    \item $Param2 = \{\tau_{GPU}:5\%,\tau_{PIN}:10\%,~\tau_{PAGE}:20\%\}$
\end{itemize}

Since these parameters do not affect the inference accuracy of RAG, this experiment focuses on comparing inference efficiency. We find that the TTFT speedup for both $Param1$ and $Param2$ is about 1.66$\times$. In summary, based on the above experimental results, we set the three threshold parameters in Algorithm~\ref{alg:heataware} to $\tau_1 = 10\%$, $\tau_2 = 10\%$, and $\tau_3 = 10\%$, and the three threshold parameters in Algorithm~\ref{alg:data_placement} to $\tau_{GPU} = 5\%$, $\tau_{PIN} = 5\%$, and $\tau_{PAGE} = 10\%$ for all subsequent experiments.

\begin{figure}
    \centering
    \subfloat[Param1]{
        \includegraphics[width=0.49\linewidth]{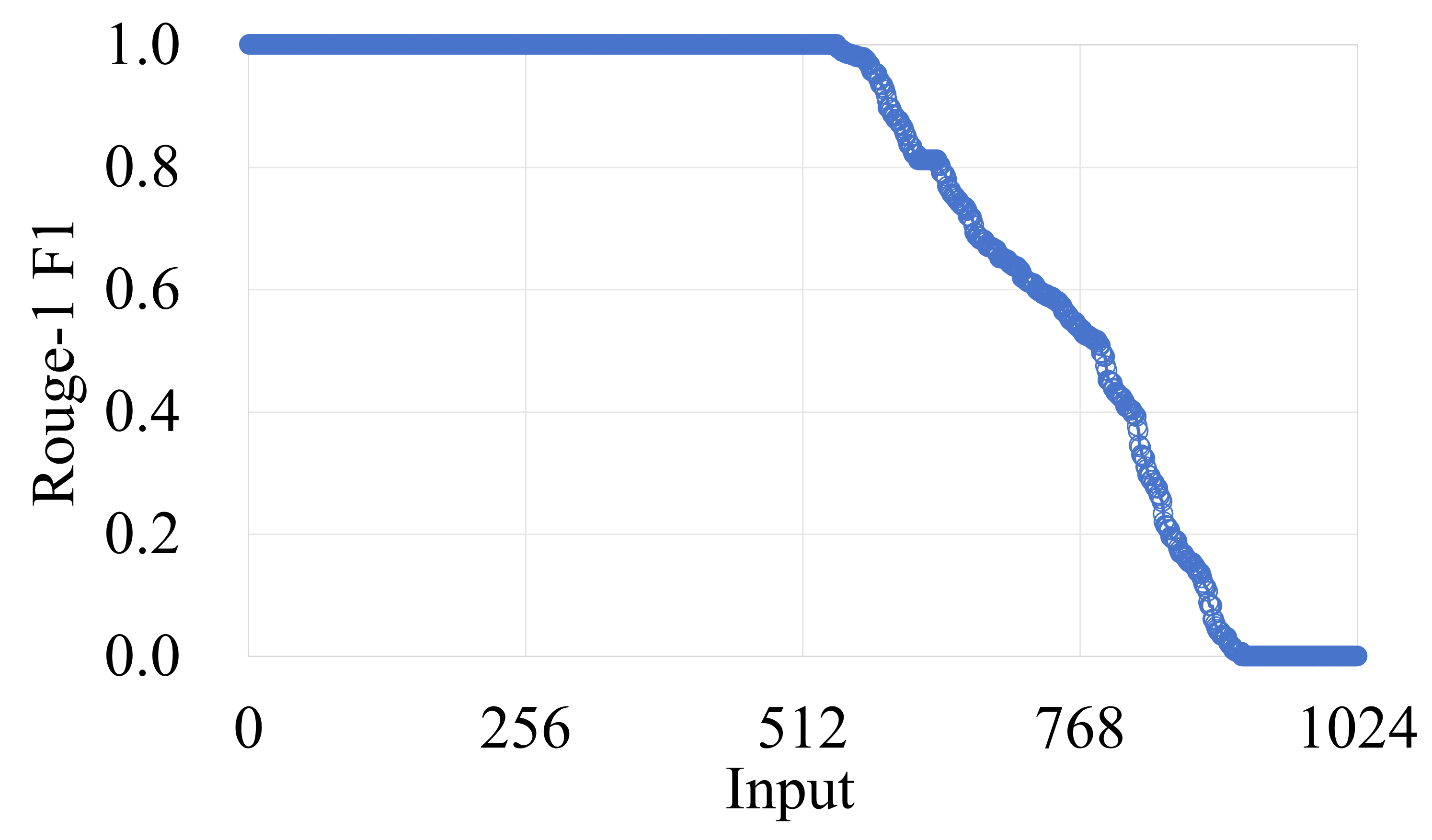}
        \label{fig:config-ROUGE-1}
    }
    \subfloat[Param2]{
        \includegraphics[width=0.49\linewidth]{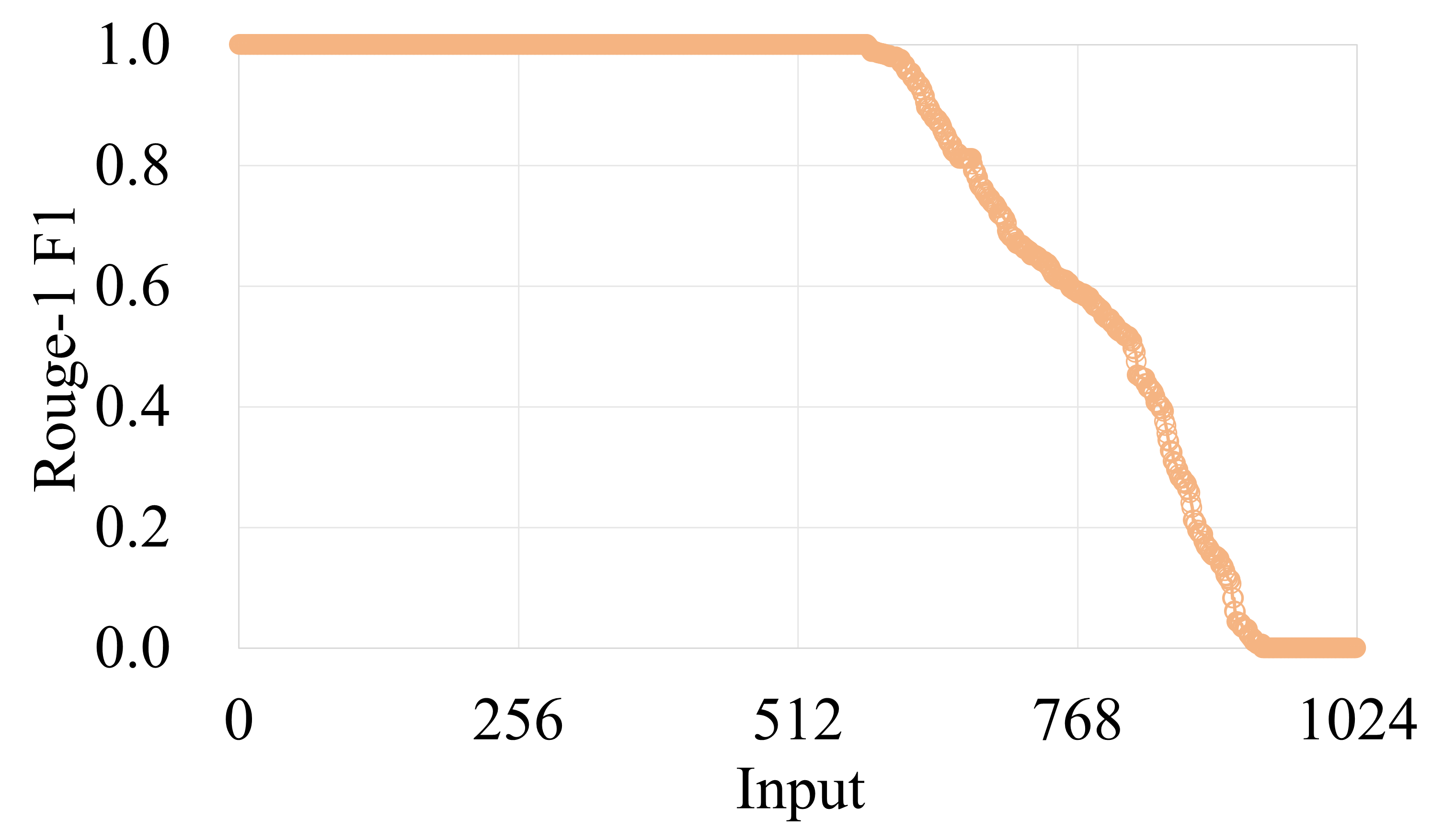}
        \label{fig:config-rouge-2}
    }
    \\
    \subfloat[Param3]{
        \includegraphics[width=0.48\linewidth]{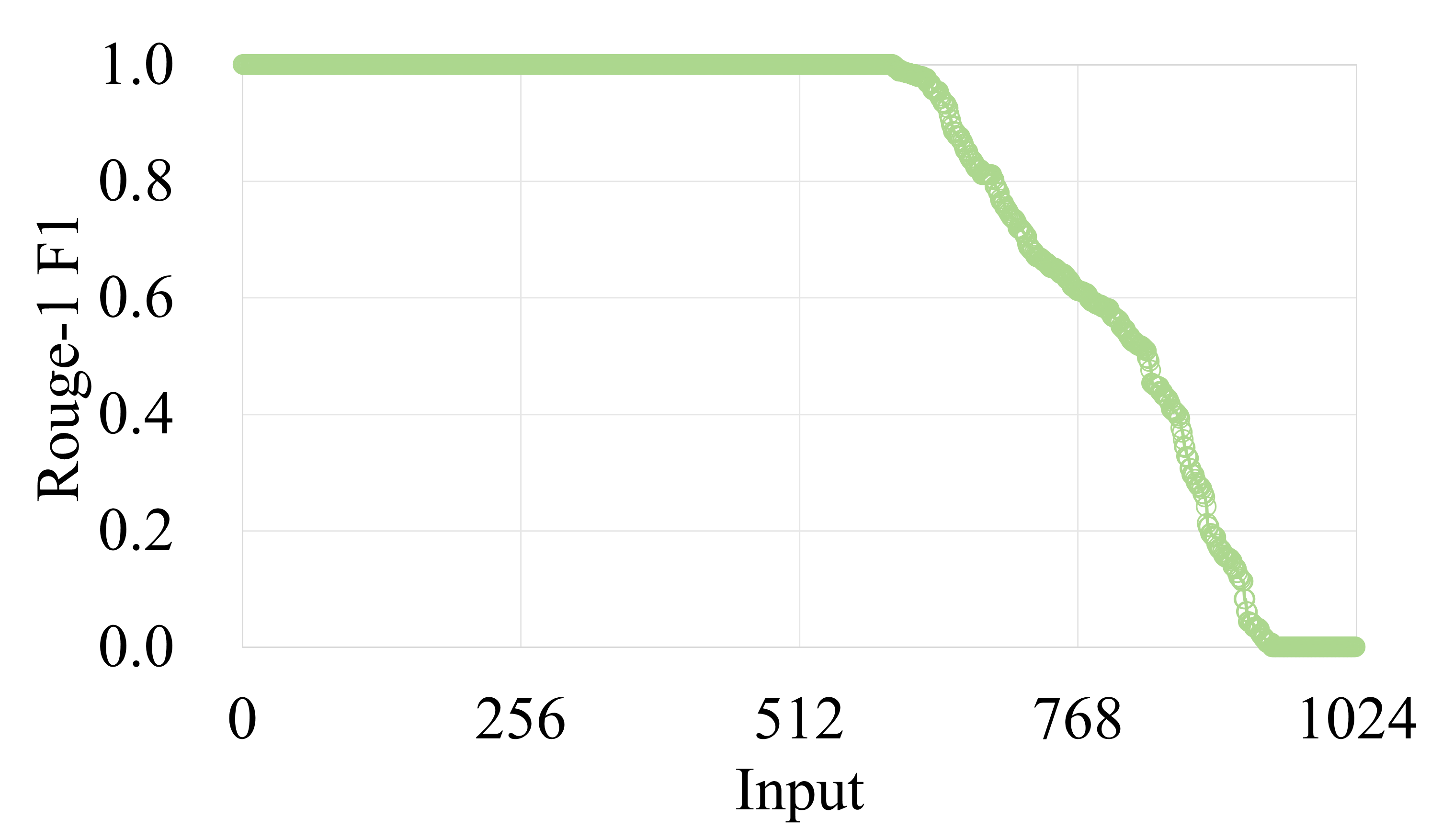}
        \label{fig:config-rouge-3}
    }
    \subfloat[Param4]{
        \includegraphics[width=0.49\linewidth]{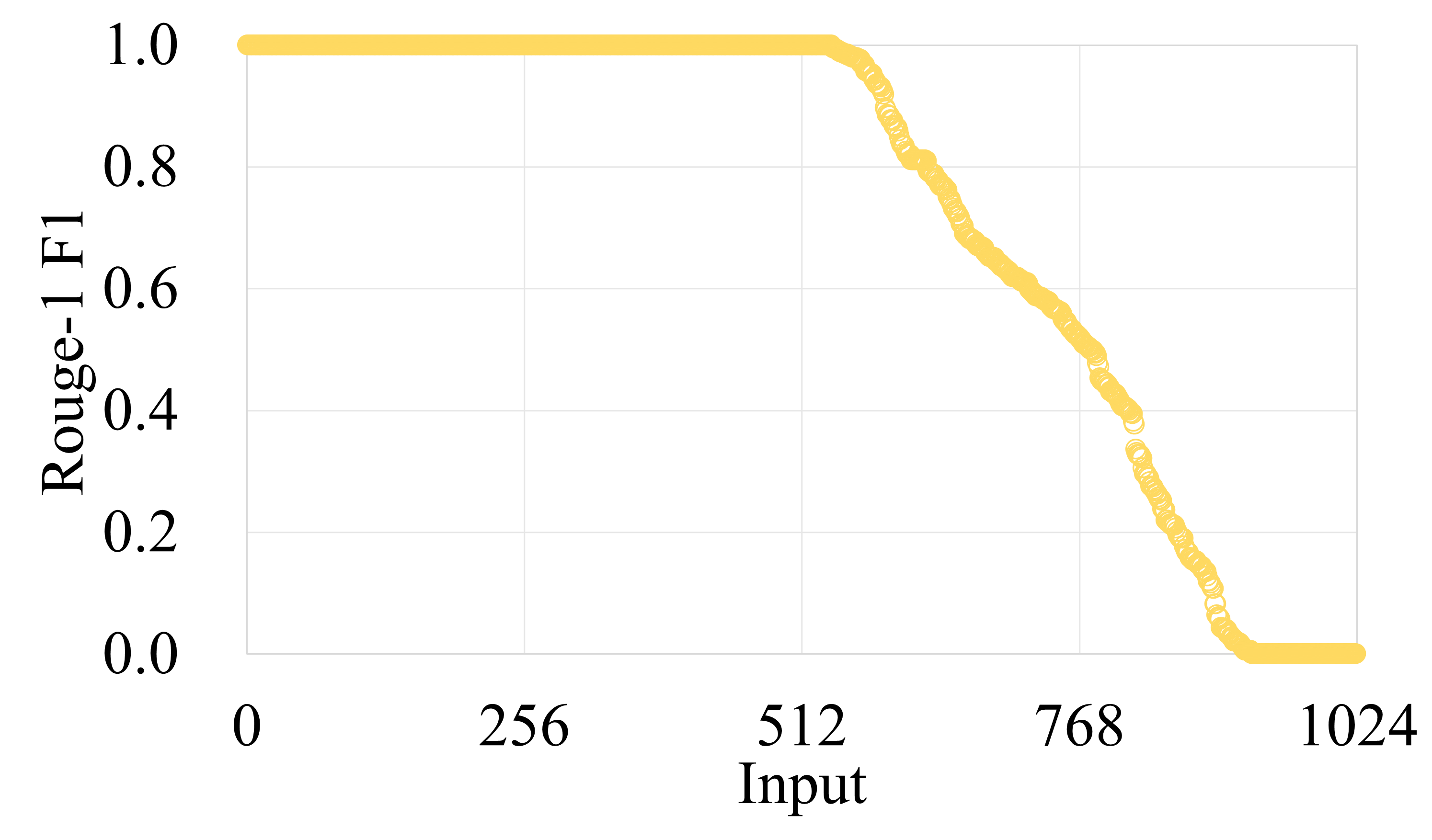}
        \label{fig:config-rouge-4}
    }
    \caption{ROUGE-1 F1 scores comparison of $\tau_1$, $\tau_2$, and $\tau_3$ under four parameter configurations.}
    \label{fig:config-ROUGE}
\end{figure}


\subsection{Comparison with the SOTA}

\begin{figure}[t]
    \centering
    \includegraphics[width=0.95\linewidth]{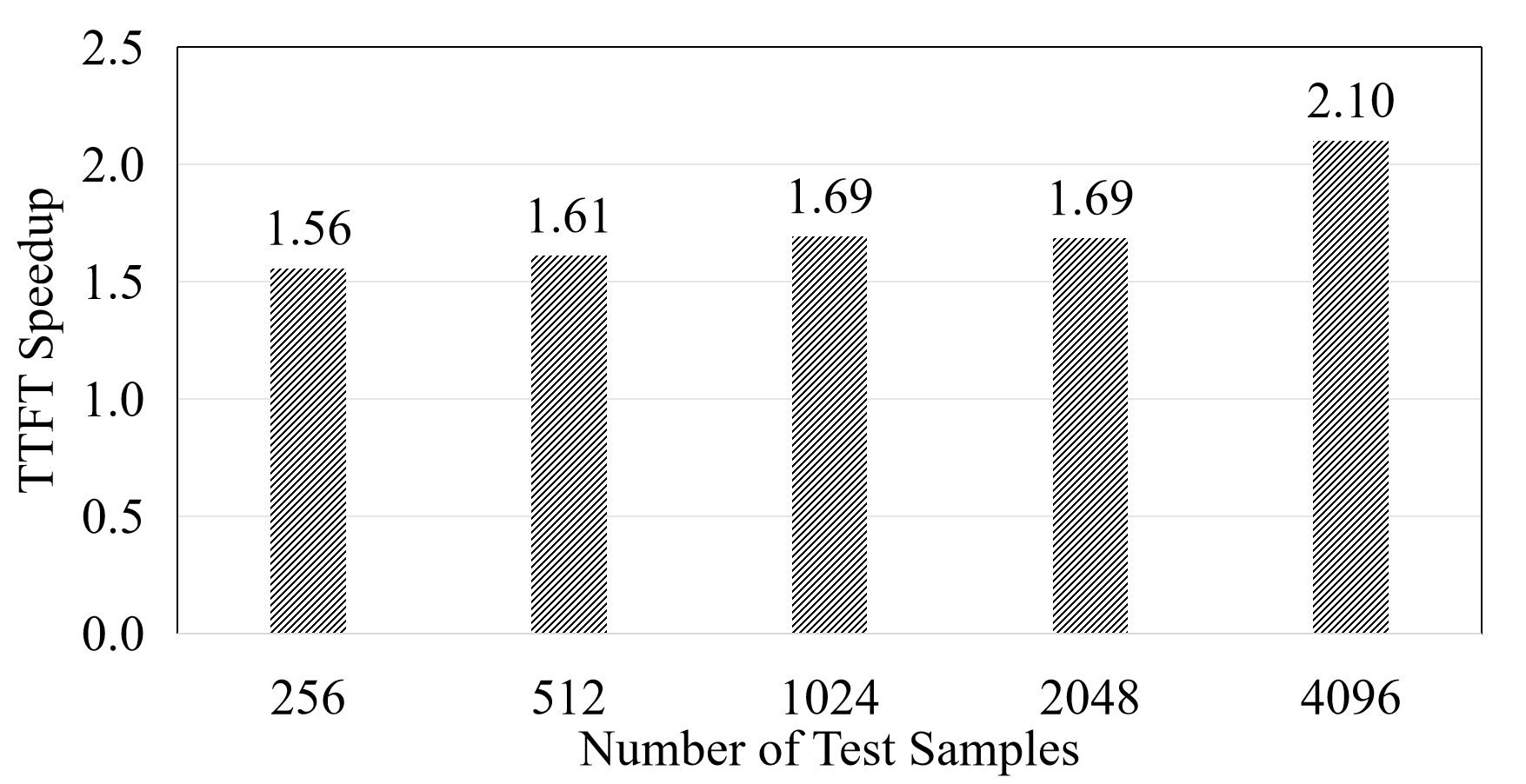}
    \caption{Experimental results of HA-RAG on datasets of varying sizes. }
    \label{contrast}
\end{figure}

In this subsection, we compare the performance and accuracy of our proposed HA-RAG with TurboRAG. The test set sizes are set to 256, 512, 1024, 2048, and 4096, respectively. Figure~\ref{contrast} shows the TTFT speedup of HA-RAG over TurboRAG under different test set sizes. It can be observed that, overall, the speedup of HA-RAG increases with the size of the test set. When the test set size reaches 4096, the average speedup achieves 2.10$\times$. This is because, as the test set size increases, the frequency of I/O operations for loading KV chunks from disk into memory gradually decreases, while the hit rates of pageable memory, pinned memory, and GPU memory increase. As a result, the acceleration effect of HA-RAG improves accordingly.

\begin{figure}[t]
    \centering
    \includegraphics[width=0.95\linewidth]{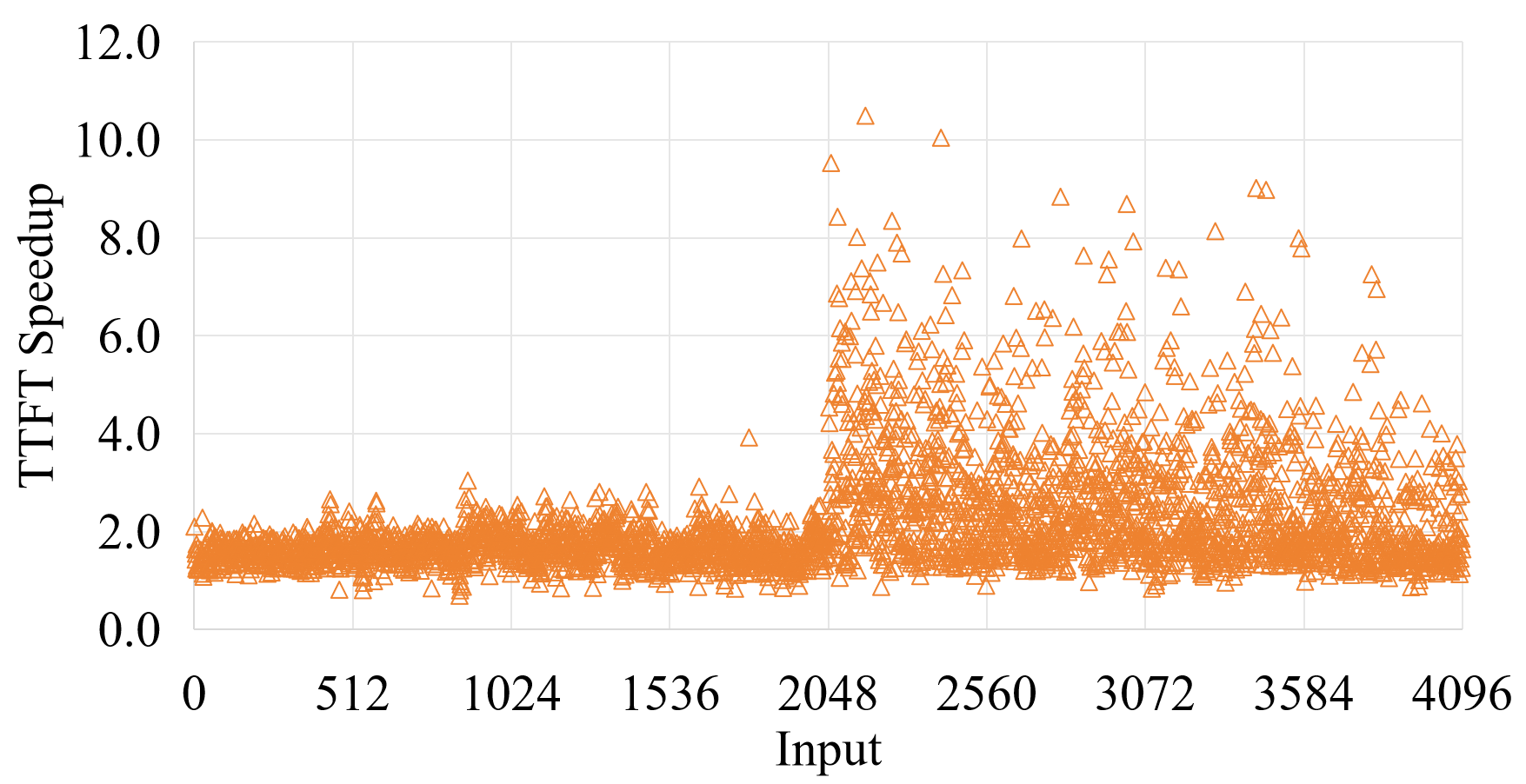}
    \caption{Speedup of HA-RAG on 4,096 inputs. }
    \label{contrast-4096TTFT}
\end{figure}

To further analyze the acceleration effect of HA-RAG, we conducted a statistical analysis of the speedup across 4,096 inputs, as shown in Figure~\ref{contrast-4096TTFT}. We observed that for the first 2048 inputs, the average speedup of HA-RAG is relatively modest, around 1.65×. As KV chunks are gradually optimized and placed across different storage, the frequency of HA-RAG loading or reading data from high-speed storage such as GPU memory and pinned memory increases during the processing of subsequent tasks, leading to improved overall efficiency. For the latter 2048 inputs, HA-RAG achieves an average speedup of 2.55$\times$, with a maximum speedup of 10.49$\times$.

The impact of the HA-RAG method on text generation quality is shown in Figure~\ref{fig:config-ROUGE}(c). For 1,024 inputs, HA-RAG is able to generate results that are identical to those of TurboRAG (ROUGE-1 F1=1) for approximately 60\% of the inputs. On more than 80\% of the inputs, HA-RAG achieves a ROUGE-1 F1 score greater than 0.5. These results indicate that HA-RAG can significantly improve inference efficiency without noticeably affecting the quality of text generation.

\subsection{Ablation Study}

This subsection evaluates the performance improvements brought by different optimization techniques proposed in this paper under a dataset size of 4096. Four experimental configurations are used:

\begin{itemize}
    \item The first configuration uses only the heat-aware mixed-precision compression and loading method, denoted as MP-only.
    \item The second configuration uses only the heat-aware data placement optimization, but without optimization involving pinned memory, denoted as DP(w/o-Pin).
    \item The third configuration uses only the data placement optimization in pinned memory, denoted as DP(Pin-only).
    \item The fourth configuration applies data placement optimization across multi-level memory without using mixed-precision compression and loading, denoted as DP-only.
\end{itemize}

\begin{figure}[t]
    \centering
    \includegraphics[width=0.95\linewidth]{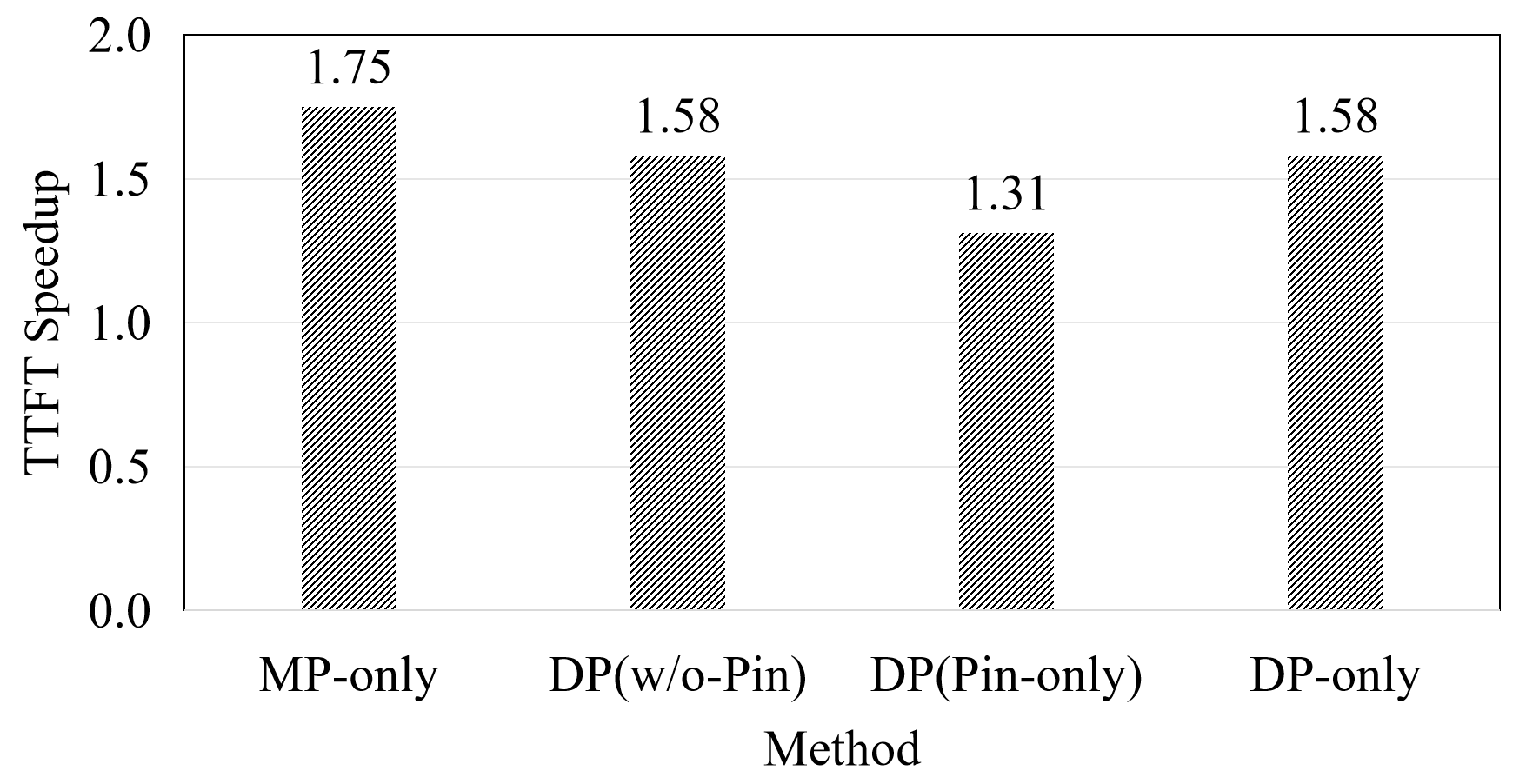}
    \caption{Ablation study results. }
    \label{ablation}
\end{figure}

Figure~\ref{ablation} shows the TTFT speedup of HA-RAG relative to TurboRAG under these four experimental configurations. From the figure, we can observe that MP-only achieves a significant performance improvement, delivering a 1.75$\times$ speedup over the baseline. The reason is that the mixed-precision compression method reduces the KV chunk size by about half, significantly lowering the overhead of KV chunk loading, memory usage, and memory access. In contrast, the performance gains from data placement optimization strategies are relatively limited. On one hand, frequently accessed KV chunks are often automatically placed in CPU pageable memory by the operating system, which usually resides in CPU RAM. Since transferring data from CPU pageable memory to pinned memory and from pinned memory to GPU memory is significantly faster than loading data from disk, the performance bottleneck remains at the initial data loading from disk. On the other hand, if a KV chunk is repeatedly accessed by adjacent prompts, it's highly likely that it already resides in GPU memory during subsequent accesses, making it difficult for data placement optimizations to deliver further speedup. 

\section{Related Work}

A considerable body of research has been devoted to improving the inference efficiency of RAG systems. These optimization techniques primarily focus on four aspects: KV cache precomputation and reuse, KV cache compression, improvements in retrieval efficiency, and scheduling strategy optimization. 

In terms of \textbf{KV cache precomputation}, several studies aim to reduce computation overhead during inference by generating KV caches for retrieved documents in advance. TurboRAG \cite{lu2024turborag} segments documents and precomputes Keys and Values for each chunk, allowing the inference phase to directly load relevant KV chunks, thus substantially reducing computational redundancy and achieving acceleration. KVLINK \cite{yang2025kvlink} further enhances this by stitching together multiple precomputed KV chunks to support diverse context requirements, thereby improving inference efficiency and scalability. Work \cite{yang2025ape} combines KV cache precomputation with an adaptive parallel encoding strategy to deliver end-to-end performance gains while maintaining generation quality.

Beyond precomputation, \textbf{KV cache reuse} is also widely employed to eliminate redundant computation and improve resource utilization. HyperRAG \cite{an2025hyperrag} leveraged historical KV caches along with system-level optimizations to improve inference throughput. CacheBlend \cite{yao2025cacheblend} selectively recomputed and updated the KV cache for certain tokens and executes these updates in parallel with the retrieval process, thereby significantly reducing the TTFT without affecting output quality. Cache-Craft \cite{agarwal2025cache} introduced a strategy to evaluate the reusability of KV chunks and employed an efficient recomputation mechanism to mitigate potential quality degradation caused by reuse. Block-attention \cite{ma2024block} partitioned retrieved documents into discrete blocks and computed KV independently, which helps reduce initial inference delay. Work \cite{couturier2025semantic} further reduced redundant KV computation by reusing context information from similar queries, thereby improving overall efficiency.

For \textbf{KV cache compression}, CacheGen \cite{liu2024cachegen} employed a custom tensor encoder that exploits the distributional properties of KV caches to encode them into compact bitstream representations, aiming to reduce both network and computational latency. This method proves particularly useful in resource-constrained or large-scale deployment scenarios.

Regarding \textbf{retrieval efficiency}, various studies propose forward-looking and structured retrieval mechanisms. CAG \cite{chan2025don} eliminated retrieval latency through preloading, simplifying system execution. RAGCache \cite{jin2024ragcache} organized intermediate states of retrieved knowledge into a tree structure and incorporates caching and speculative inference, enabling parallel execution of retrieval and generation and significantly boosting efficiency. Proximity \cite{bergman2025leveraging} reduced high-latency retrieval operations by adopting an approximate KV cache mechanism. TELERAG \cite{lin2025telerag} introduced a lookahead retrieval mechanism coupled with asynchronous data transfer to improve end-to-end latency. AquaPipe \cite{yu2025aquapipe} applied prefetching strategies to parallelize retrieval and inference, increasing overall throughput. LightRAG \cite{guo2024lightrag} introduced graph-based structures and a dual-layer indexing mechanism to enhance retrieval accuracy and efficiency simultaneously.

In the realm of \textbf{scheduling strategies}, researchers have explored a variety of resource management and execution models to alleviate bottlenecks. VectorLiteRAG \cite{kim2025adaptive} dynamically placed frequently accessed vector indices into high-bandwidth memory (HBM), improving system responsiveness. Cake \cite{jin2024compute} proposed a bidirectional and adaptive scheduling mechanism to enable parallel computation of multiple tasks, thereby enhancing overall system throughput. RAGDoll \cite{yu2025ragdoll} adopted a pipelined parallelism approach to concurrently process retrieval and inference, effectively reducing latency. Work \cite{luo2025does} developed a specialized retriever for precise retrieval of critical pages and employs unsupervised post-training to optimize the model’s ability to utilize the retrieved information.

In addition to these mainstream directions, several other studies present novel optimization strategies from the perspective of \textbf{system architecture and cache resource management}. By encoding the retrieved documents in parallel, work \cite{zhu2024accelerating} eliminated the latency introduced by long-range attention over the retrieved content. Work \cite{lee2025shared} introduced a disk-based shared KV cache management system that improves throughput and latency under constrained resource conditions. KVShare \cite{bergman2025leveraging} utilized a Dual-Stage High Deviation (DHD) mechanism and a cache-aware scheduler to improve throughput. Task-KV \cite{he2025task} dynamically allocates KV cache resources based on task types, enhancing system flexibility. RAGServe \cite{ray2024ragserve} reduced inference latency by tuning query configuration parameters while maintaining generation quality. HCache \cite{gao2025fast} proposed restoring LLM states from intermediate activations to improve context processing efficiency. Work \cite{DBLP:conf/sigcse/YuL0BL25} proposed a novel RAG framework-Intelligence Concentration (IC)-which mitigates the challenge in RAG systems of simultaneously achieving high accuracy, short context length, and low response time. DRAGON \cite{couturier2025semantic} enhanced generation quality by integrating general and personalized knowledge while employing an adaptive scheduling algorithm in a distributed setting, reducing TTFT and preserving privacy.

Collectively, these methods significantly enhance system throughput and accelerate inference speed. However, these studies pay limited attention to the access patterns and data distribution characteristics of KV caches within retrieved documents, leaving room for further refinement in this aspect. Building upon this foundation, our work further explores the optimization of RAG inference efficiency by leveraging the observation on access patterns and data distribution KV chunks, thereby pushing the performance boundary of retrieval-augmented generation systems.

\section{Conclusion}
This paper addresses the critical KV chunks loading and memory access bottlenecks in RAG systems through a novel hotness-aware optimization framework. By analyzing the access patterns and numerical distributions of KV chunks, we develop an approach combining mixed-precision compression with hotness-aware data placement optimization. Our solution employs four precision-reduced representations: INT8, FP8 formats (E4M3 and E5M2), and an enhanced grouped shared-exponent method, to achieve mixed-precision compression for KV chunks based on their access frequency. Furthermore, we introduce a hotness-aware data placement optimization method that prioritizes high-frequency chunks in faster memory hierarchies. Comprehensive experiments demonstrate that HA-RAG achieves substantial performance gains over TurboRAG, delivering an average speedup of 2.10x while introducing negligible accuracy degradation. These advancements significantly improve the practical deployment efficiency of RAG systems, establishing an effective balance between inference performance and accuracy. 

\bibliographystyle{IEEEtranS}
\bibliography{refs}

\end{document}